%% file: main.tex
\documentclass[conference]{IEEEtran}
\usepackage{times}

\usepackage[numbers]{natbib}
\usepackage{multicol}
\usepackage[bookmarks=true]{hyperref}
\usepackage{amsfonts}
\usepackage{xcolor}
\usepackage{graphicx}
\usepackage{amsmath}
\usepackage[capitalize,noabbrev]{cleveref}
\usepackage{wrapfig}
\usepackage{tikz}
\usepackage{algorithm}
\usepackage{algpseudocode}
\usepackage{array}
\usepackage{tcolorbox}
\usepackage{float}
\usepackage{booktabs}
\usepackage{multirow}
\usepackage{caption}
\usepackage{subcaption}
\usepackage{etoolbox}
\usepackage{selectp} 
\usepackage{etoc}

\usetikzlibrary{arrows.meta, positioning}

\definecolor{zoharcolor}{RGB}{111,45,168}    
\definecolor{avivcolor}{RGB}{214,72,137}     
\definecolor{elicolor}{RGB}{204,102,0}       
\definecolor{talcolor}{RGB}{220,60,30}       
\definecolor{roycolor}{RGB}{0,51,102}        
\definecolor{aloncolor}{RGB}{0,128,0}        
\definecolor{danielcolor}{RGB}{0,102,102}    

\newcommand{\refapp}[1]{Appendix \ref{#1}}
\newcommand\norm[1]{\left\lVert#1\right\rVert}

\input{defs}
\IEEEoverridecommandlockouts

\begin{document}

\title{More with LESS -- Local Scene Representations for Tactile Imaging}


\author{Zohar Rimon$^{\ast, 1}$ \thanks{$\ast$ Correspondence to zohar.rimon@campus.technion.ac.il.

Website: \href{ https://zoharri.github.io/LESS}{ \textit{zoharri.github.io/LESS}} 

Code: \href{ https://github.com/zoharri/LESS}{ \textit{github.com/zoharri/LESS/}} 

Data: \href{ https://zenodo.org/communities/artificial-palpation}{ \textit{zenodo.org/communities/artificial-palpation}}} \quad Elisei Shafer$^1$
\quad Tal Tepper$^1$
\quad Daniel Kozin$^1$ \quad Alon Malka$^1$ \\ Roy Holland$^{1, 2}$ \quad Aviv Tamar$^1$\\[0.3cm]

$^1$Technion - Israel Institute of Technology \quad $^2$Rambam Health Care Campus 
}

 \makeatletter
\let\@oldmaketitle\@maketitle
    \renewcommand{\@maketitle}{\@oldmaketitle
    \centering
    \vspace{0em}
    \includegraphics[trim=0 0 1cm 0, clip,width=0.99\textwidth]{images/cover_fig/Cover_RSS_LESS_new.png}
\captionof{figure}{\textbf{Overview of our approach:} \textit{(a)} Using an automated robotic data collection setup, we collect the largest soft-body tactile interaction dataset to date, coupled with MRI imaging of the inner structure of our modular-designed phantoms. \textit{(b)} We propose the \name\ neural network architecture to effectively represent a complex tactile scene by capturing local tactile interaction sequences, and generate local visualizations of the sensed objects. \textit{(c)} Our approach leads to zero-shot compositional generalization and maintains accurate predictions on unseen phantom configurations with multiple inclusions and different shapes. \textit{(d)} Finally, we show a proof-of-concept real-time hand-held tactile imaging system that visualizes the inner structure of soft objects.}
    \vspace{-0.2cm} \label{fig:teaser}
    \setcounter{figure}{1}
}
\makeatother

\maketitle

\input{sections/abstract}
\IEEEpeerreviewmaketitle

\input{sections/introduction}

\input{sections/related_works}

\input{sections/formulation}

\input{sections/local_representation_architecture}

\input{sections/experiments}

\input{sections/conclusion}

\input{sections/acknowledgments}

\newpage
\bibliographystyle{plainnat}
\bibliography{references}

\input{sections/appendix/appendix}

\end{document}

%% file: defs.tex
\newcommand{\spose}{x}
\newcommand{\recfield}{\beta}
\newcommand{\lpose}{\bar{x}}
\newcommand{\lossfull}{\ell}
\newcommand{\lossparticle}{\bar{\ell}}
\newcommand{\controlfun}{g}
\newcommand{\history}{h}
\newcommand{\distance}{D}
\newcommand{\patch}{p}
\newcommand{\patchres}{N_p}
\newcommand{\patchlen}{\rho}
\newcommand{\patchresz}{N_z}
\newcommand{\imclasses}{\kappa}
\newcommand{\estpose}{\hat{x}}
\newcommand{\ttime}{t}
\newcommand{\softbody}{M}
\newcommand{\bodyobs}{I}
\newcommand{\force}{f}
\newcommand{\forcepred}{\hat{f}}
\newcommand{\forcedim}{k}
\newcommand{\trajlength}{T}

\newcommand{\repr}{z}
\newcommand{\lrepr}{\bar{z}}
\newcommand{\nparticles}{N}

\newcommand{\reprdim}{d_z}
\newcommand{\subsample}{K}
\newcommand{\name}{LESS}
\newcommand{\acronym}{\textbf{L}ocal \textbf{E}ncoder for \textbf{S}patial \textbf{S}ensing}

%% file: sections/abstract.tex
\begin{abstract} \label{section:abstract}
Tactile imaging seeks to reconstruct the internal structure of soft objects through touch sensing, with applications in medical diagnosis and robotic manipulation. Recent self-supervised learning approaches have shown promising results, but rely on global, unstructured representations and robot-controlled sensing, limiting generalization and practical use. We propose \acronym\ (\name), an object-centric tactile representation that exploits the local nature of touch. The tactile scene is modeled as a grid of recurrent encoders with local receptive fields, whose states are fused to reconstruct 2D or 3D images of internal structure. This compositional design enables strong generalization: models trained on single-inclusion phantoms accurately image objects with multiple inclusions and varying sizes. The local structure further supports spatial uncertainty estimation. In addition, we enable hand-held tactile imaging via external pose tracking and human-like palpation data, and extend tactile imaging to full 3D reconstruction. 
\end{abstract}

%% file: sections/introduction.tex
\section{Introduction} \label{section:intro}

Humans have a remarkable capability to understand and visualize the world through touch. Tactile scene understanding is important for medical tactile imaging (TI, \cite{egorov2008mechanical}) and for unlocking robotic capabilities in medical, domestic, and agricultural domains. Although tactile sensor hardware has advanced recently~\cite{lambeta2020digit,lee2020nanomesh,xelauskin,du2023design}, scene understanding from tactile data 
is still a challenging algorithmic problem, especially when non-rigid objects are involved. 

The recent work of \citet{Rimon2025ArtificialPalpation} proposed a self-supervised learning (SSL) based approach to TI. In their work, a manipulator equipped with a tactile sensor automatically palpates a set of soft objects (breast phantoms), for which ground truth internal structure is available through magnetic resonance imaging (MRI). A tactile representation is learned from the SSL data by training an encoder-decoder model. The encoder processes a sequence of tactile measurements and sensor positions into a ``representation'' vector, while the decoder predicts from the representation the tactile measurement at a future position. Following the SSL training, an image generation model is trained to predict internal structure (MRI slice) from the tactile representation. 
\citet{Rimon2025ArtificialPalpation} showed that with sufficient SSL training, the network can learn complex patterns in the tactile data to yield accurate imaging. 

However, several features in the method of \citet{Rimon2025ArtificialPalpation} limit its practical applicability for tactile imaging. The first is the representation. Representing a tactile scene using a single \textit{unstructured} vector ignores the compositional nature of physical objects, and requires training on ``all possible object combinations'' to generalize, which may be too costly.
Our insight, similar to object-centric representations in computer vision~\cite{locatello2020object,daniel2022dlp}, is that touch is inherently local: the tactile response of the internal structure of an object at some position is \textit{nearly independent} of the internal structure at far away positions. Based on this idea, we propose \acronym\ (\name) -- a novel object-centric representation for tactile scenes. The \name\ encoder is composed of a grid of `particles'\footnote{In this work, we use the term particle to describe a neural network representation that is also situated in space, inspired by the object-centric representation literature~\cite{daniel2023ddlp}. This term should not be confused with particle filters from the state estimation literature. }, where each particle is a recurrent neural network that processes the sequence of tactile measurements within its spatial receptive field. The \name\ image generator combines the latent state from all the particles into a single 2D or 3D image of the scene.
We show that by training on phantoms with single inclusions as in \cite{Rimon2025ArtificialPalpation}, we can produce accurate tactile images for objects with multiple inclusions at test time. Moreover, we demonstrate that \name\ yields accurate imaging for objects of different size from training. Finally, we show that the local structure of \name\ allows us to visualize the \textit{uncertainty} of the tactile scene in different areas, which may be used to direct the operator to collect additional measurements.

Another shortcoming in the method of \citet{Rimon2025ArtificialPalpation} is that it must be robot-operated at test time, limiting its use as a hand-held tactile imaging device. The reason is twofold: the method requires accurate pose tracking of the sensor obtained by the robot kinematics, and the training data contains only vertical movements, which are very different from natural human palpation. We relax these limitations by adding external position tracking of the sensor, and collecting robot palpation data that more resembles human motion. These technical modifications allow us to demonstrate the first hand-held tactile imaging device based on SSL. Finally, the imaging in \cite{Rimon2025ArtificialPalpation} was limited to 2D. We extend the method to 3D imaging, and interestingly find that 3D reconstruction also improves upon the results in \cite{Rimon2025ArtificialPalpation} when considering 2D image slices from our predicted 3D volume.

%% file: sections/related_works.tex
\section{Related Works} \label{section:related}

\textbf{Tactile imaging:} Mechanical/tactile imaging systems formalized pressure–position sensing and inversion to characterize tissue stiffness and lesion geometry in vivo \cite{wellman2001tactile,egorov2008mechanical,egorov2006prostate,weiss2008pmi,egorov2010vti}. Comprehensive reviews position tactile imaging as a low-cost elastography modality~\cite{sarvazyan2011overview} and outline requirements for broader diagnostic adoption \cite{sarvazyan2011mi,hampson2022ti}. However, state-of-the-art methods are based on averaging readings of tactile sensor arrays to generate a force map, and cannot detect subtle tissue features that are apparent to human touch and important for diagnosis~\cite{egorov2008mechanical, rimon2022meta}. In breast imaging, Hampson et al.~\cite{hampson2022ti} highlight the inability to detect background tissue elasticity, lesion position within the breast, and lesion acutance and mobility -- capabilities that an SSL approach can potentially acquire with enough training data. A related line of work explored robotic palpation \cite{khanna2024robotics,jenkinson2023robotic,syrymova2025breast,scimeca2022action,sanni2022deep}, while we only use a robot for data collection; as we demonstrate, our artificial palpation device can be hand-held.

\textbf{Representations for tactile data:} Neural networks were used for various tactile processing tasks, including material classification \cite{gao2016tactile,yuan2018active,luo2018vitac,andrussow2023minsight}, grasp-success prediction \cite{calandra2017feeling}, slip detection \cite{li2018slip}, and contour-following control \cite{lepora2019pixels}. Representation learning for tactile data is an emerging area. 

Several studies investigated learning representations for image-based tactile sensors. Sparsh~\cite{higuera2024sparsh} is an SSL representation for single tactile measurements, and AnyTouch~\cite{feng2025anytouch} is an SSL representation for short ($3$-frame) tactile videos. For tactile sensor arrays, T-DEX~\cite{guzey2023dexterity} is an SSL representation for a single measurement of a taxel grid. All of these methods learn to represent individual tactile measurements, while our work is on representing tactile \textit{scenes}.

For tactile scene representations, prior work excluding \citet{Rimon2025ArtificialPalpation} focused almost exclusively on rigid objects. \citet{qi2023general} learned a representation for a sequence of tactile, proprioceptive, and visual observations, using simulated 3D point-clouds of objects during training. NeuralFeels \cite{suresh2024neuralfeels} uses neural fields to represent the 3D shape of an object from vision and tactile sensors, Tactile-Informed 3DGS~\cite{comi2025snap} employ Gaussian splatting for similarly structured data, and VTacO~\cite{xu2023visual} predict the surface of an object from vision and touch by learning a winding number field. None of these methods can be used for imaging the inner structure of soft objects, which is critical for medical applications and our focus here.

%% file: sections/formulation.tex
\section{Formulation} \label{section:formulation}

We follow the artificial palpation problem formulation of \citet{Rimon2025ArtificialPalpation}. 
A rigid tactile sensor is controlled to be at time $\ttime \in 0,\dots,\trajlength$, in pose $\spose_\ttime \in \mathbb{R}^{6}$. 
The sensor touches a soft body $\softbody$, and produces a $\forcedim$-dimensional force reading $\force_\ttime \in \mathbb{R}^{\forcedim}$. Here, $\softbody$ is generally unknown and represents all the structural and mechanical properties of the body that determine the force on the sensor. Subsequently, the sensor is moved to the next pose $\spose_{\ttime+1}$ by a controller, and the next reading is obtained. In addition, we may have an observation of the soft body, denoted $\bodyobs$, for example, an MRI scan. 

We make two modifications to the formulation in \cite{Rimon2025ArtificialPalpation}. The first is sensor pose estimation: we cannot observe $\spose_\ttime$, but only a noisy estimate $\estpose_\ttime$. The second is the controller. While \cite{Rimon2025ArtificialPalpation} considered a fixed motion for the sensor, we investigate both different controllers and human-operated hand-held control. We denote $\spose_{\ttime+1} = \controlfun\left(\history_t\right)$ a general control function that  depends on $\history_t$, the history of measurements up to time $t$.

%% file: sections/local_representation_architecture.tex
\section{Method} 

This section describes our method for learning tactile imaging. We begin with the \name\ architecture, continue with training details and data collection, and finally describe technical contributions required for hand-held tactile imaging.

\begin{figure*}[t]
    \centering
    \includegraphics[width=.9\textwidth]{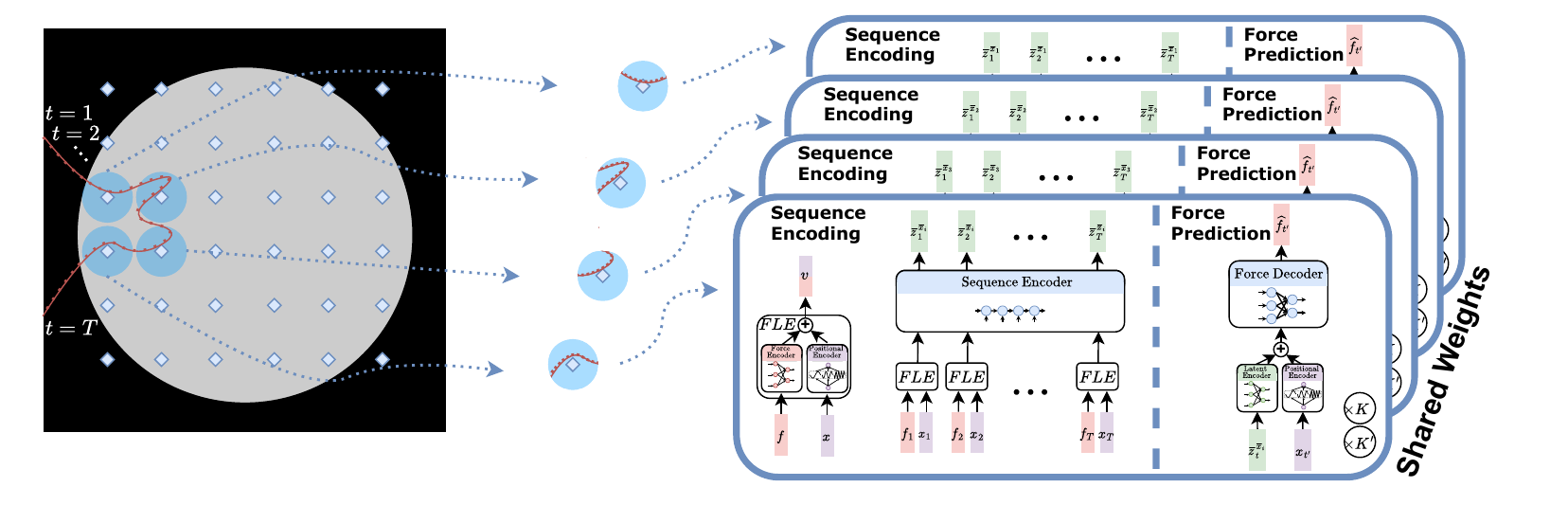}
    \caption{\textbf{\name\ representation learning:} A set of localized representations for processing a tactile scene, where each representation attends only to measurements within its receptive field. For a localized representation positioned at $\lpose$, the sequence of tactile data within its receptive field is centered around $\lpose$, and processed with a GRU-based encoder-decoder architecture, with shared weights. The encoder-decoder figure on the right is modified from \cite{Rimon2025ArtificialPalpation}. 
   }
    \label{fig:LESS_rep}
\end{figure*}

\subsection{\name\ - \acronym}
\label{section:architecture}

The method of \cite{Rimon2025ArtificialPalpation} proposed a recurrent neural network that encodes a sequence of tactile measurements $\left\{\spose_0,\force_0,\dots, \spose_\ttime,\force_\ttime \right\}$ into a $\reprdim$-dimensional latent vector representation $\repr_t \in \mathbb{R}^{\reprdim}$. Our main insight is that tactile scenes, where material properties at a particular position $\spose$ typically do not affect the force sensed at a distant position $\spose'$, can be better represented by a collection of $\nparticles$ local representations $\lrepr_t = \left\{ \lrepr^{\lpose_1}_t, \dots, \lrepr^{\lpose_{\nparticles}}_t\right\}$, where each local representation $\lrepr^{\lpose_{i}}_t$ has a \textit{spatial receptive field} centred around position $\lpose_{i}$. The motivation for such a disentangled representation is two-fold, as we describe next.

\textbf{Compositional Generalization:} Consider objects that are composed of several different material properties at different positions. A single vector representation must learn to represent \textit{all the combinations} of different materials at different positions, the quantity of which grows exponentially with the number of materials. Our disentangled representation only needs to learn the local tactile response for the possible materials, while the compositional generalization is by construction.  

\textbf{Localized Uncertainty Estimates:} For a hand-held device, it is important to convey to the operator areas with high uncertainty, to guide data collection at test time. We hypothesize that localizing the uncertainty using a disentangled representation is easier than using a single vector representation, where uncertainty at multiple positions may be entangled.

\begin{figure}[t]
    \centering
    \includegraphics[trim=2cm 0 0 0, clip,width=.85\linewidth]{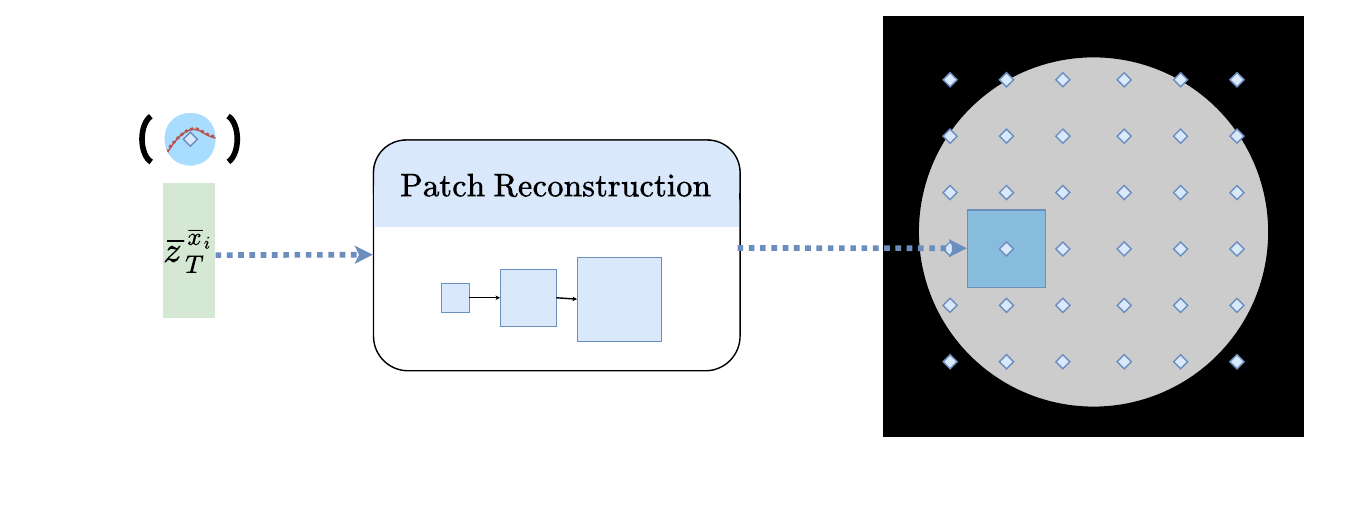}
    \caption{\textbf{\name\ image reconstruction:} Each learned local representation is mapped to a local image patch using a transposed convolution neural network. In turn, all patches are stitched together to produce the final image by averaging logits and applying a softmax.  
   }
    \label{fig:LESS_rec}
\end{figure}

Our architecture, \acronym\ (\name), is composed of a tactile sequence encoder-decoder module, and an image reconstruction module, depicted in Figure \ref{fig:LESS_rep} and Figure \ref{fig:LESS_rec}, and described in the following.

\subsubsection{\name\ Sequence Encoder}
The local representation $\lrepr^{\lpose_{i}}_t$ is obtained from the input sequence $\left\{\spose_0,\force_0,\dots, \spose_\ttime,\force_\ttime \right\}$ by applying the following sequence of steps.

\textbf{Receptive Field:} Each $\lpose_{i}$ has a receptive field $\mathcal{X}_i = \left\{\spose: \distance(\spose, \lpose_{i}) < \recfield\right\}$, for some distance function $\distance$. We filter out inputs that are outside the receptive field of $\lpose_{i}$, that is, only pairs $\spose_j,\force_j$ that satisfy $\spose_j \in \mathcal{X}_i$ are kept in the input sequence to $\lpose_{i}$. 

\textbf{Centering:} Every input pose $\spose_j$ in the input sequence to $\lpose_{i}$ is centered around $\lpose_{i}$ by subtracting $\lpose_{i}$ from it. 

\textbf{Force-Location Encoder (FLE):} We follow \cite{Rimon2025ArtificialPalpation} and encode each input pair $\spose_j,\force_j$ by first encoding the centered $\spose_j$ using sinusoidal positional encoding (PE), then applying an MLP on $\force_j$, and finally summing the MLP and PE outputs.

\textbf{Recurrent Neural Network:} The sequence of FLE encoded forces and poses is input to a Gated Recurrent Unit (GRU, \cite{gru}), and $\lrepr^{\lpose_{i}}_t$ is obtained as the GRU hidden state. 

In our implementation, the $\nparticles$ GRUs for different positions have shared weights, as relevant patterns in localized sensory readings from different positions should be similar. 

In addition, we positioned the $\lpose_{i}$'s on a uniform $2$-dimensional grid, and our distance function was Euclidean on the 2-dimensional $x-y$ plane, namely, $\distance(a, b) = \sqrt{(a_x-b_x)^2 + (a_y - b_y)^2}$. This choice is motivated by the fact that our objects mostly vary in the planar position of different inclusions. More general settings may require a $3$-dimensional grid, and possibly accounting for orientations.

\textbf{Force Decoder:} To predict $\forcepred(\spose, \lrepr^{\lpose_{i}})$, the force at position $\spose$ from a local representation $\lrepr^{\lpose_{i}}$, we first center $\spose$ around $\lpose_{i}$, as described above, then apply PE to the centered $\spose$ and an MLP to $\lrepr^{\lpose_{i}}$, and finally input the sum of the PE and MLP to another MLP that predicts the force.

\subsubsection{\name\ Image Reconstruction}
Given the encoded representation, which essentially captures information about the palpated object, we seek to produce an image of the object's internal structure. In \cite{Rimon2025ArtificialPalpation}, this was done by mapping the single representation vector $\repr_t$ to a $2$-dimensional image using transposed convolutions.

In contrast, we seek to map the set of local representations $\left\{ \lrepr^{\lpose_1}_t, \dots, \lrepr^{\lpose_{\nparticles}}_t\right\}$ to an image such that each reconstructed pixel is only affected by representations in its local area. In addition, we seek to image $3$-dimensional volumes, to capture a richer and more precise visualization of objects' internal structure. We begin by describing our approach to $2$-dimensional images, and later extend to $3$D.

Consider a square image patch centered on the $x-y$ components of $\lpose$, with side length $2\patchlen$. We assume that each patch contains $\patchres \times \patchres$ pixels, and that each pixel can take $\imclasses$ possible values; in our experiments, $\imclasses=3$, corresponding to free space, hydrogel, or silicone inclusion. Let $\patch(\lpose) \in \imclasses^{\patchres \times \patchres}$ denote the logits for every pixel in $\lpose$'s image patch. The logits for each patch are obtained using a transposed convolution with input $\lrepr^{\lpose}$, where the transposed convolutions for all patches have shared weights. To compose the patches into a complete image, we first sum all logits and then apply a softmax to obtain class probabilities for every pixel.

\textbf{$3$-dimensional reconstruction} The $3$-dimensional reconstruction module is similar, with the difference that each patch is a square column with $\patchres \times \patchres \times \patchresz$ pixels.

\textbf{Related Spatio-Temporal Architectures:} The \name\ architecture draws inspiration from related spatio-temporal representations in the literature, such as the convolutional LSTM~\cite{shi2015convlstm} and Deep Latent Particles~\cite{daniel2023ddlp}. Several properties specialized for tactile imaging differentiate it from previous work. First, different from convolutional neural networks, \name\ particles apply a hard threshold outside their receptive fields, which relates to the physical locality of touch. Second, \name\ decodes both forces and $2$D/$3$D images.

\subsection{SSL Training and Data}
\label{section:ssl_train_data}

Following \cite{Rimon2025ArtificialPalpation}, force prediction drives SSL of the \name\ encoder-decoder. However, we extend the force prediction method of \cite{Rimon2025ArtificialPalpation} to a localized setting. 

Our training loss $\lossfull = \frac{1}{\nparticles}\sum_{i=1}^{\nparticles} \lossparticle({\lpose}_i)$ averages the loss functions for individual particles, which we describe next. Consider a local representation centered on  ${\lpose_{i}}$, and a sequence $\left\{\spose_0,\force_0,\dots, \spose_T,\force_T \right\}$ in the data. We first filter out pose-force pairs outside $\mathcal{X}_i$, and calculate the local representations $\{\lrepr^{\lpose_i}_t\}$ as described above for all time steps in the filtered sequence. The mean-squared error (MSE) in predicting a force at position $\spose_{t'}$ given the representation $\{\lrepr^{\lpose_i}_t\}$ is $\| \forcepred(\spose_{t'}, \lrepr^{\lpose_{i}}_t) - \force_{t'}\|^2$. The loss $\lossparticle({\lpose}_i)$ averages the MSE over a set of time index pairs $t,t'$ sampled similarly to \cite{Rimon2025ArtificialPalpation} (full details in \refapp{sec:app_training_details}).

SSL data consists of force and pose sequences obtained from palpating soft phantoms in the training set. 
We collect a similar dataset to the Artificial Palpation Poke Dataset \citep{rimon2025datav1}, which was also used in \cite{Rimon2025ArtificialPalpation}, and henceforth termed \texttt{data-v1}. The sequences in this data correspond to vertical motions of the tactile sensor in a fixed orientation, touching artificial breast phantoms made from soft silicone skin and hydrogel filling, with a single, round, soft-silicone inclusion. We hypothesize that hand-held motion would be significantly different, and therefore produced two new training datasets, \texttt{data-poke} and \texttt{data-primitive}, and a small test dataset \texttt{data-handheld}. Our data generation method differs from \cite{Rimon2025ArtificialPalpation} in object fabrication method, robot palpation motion, and the automatic data collection mechanism, as we describe next; full details and reproduction steps are in \refapp{sec:app_manufacturing_details}. All datasets are publicly available.

\textbf{Phantom fabrication:} Our fabrication technique is based on curing soft silicone in 3D printed moulds, which enables accurate thickness control over the phantom `skin', and inclusions with complex shapes, such as several connected ball-shaped objects. Following the high-level design in \cite{Rimon2025ArtificialPalpation} of an outer shell and an insert with an inclusion that can be rotated in $8$ angles, we generated $40$ inserts ($24$ of which have been proposed in \cite{Rimon2025ArtificialPalpation}) and 8 shells, with inclusions varying in size and shape. See \refapp{sec:app_manufacturing_details} for details.

\textbf{Tactile Sensor:}
We use an off-the-shelf Xela uSCu ALHA sensor \cite{xelauskin},  capturing data at $100$Hz. We
emphasize that our learning method is not sensor-specific.

\textbf{Sensor Motion:} \texttt{data-poke} was collected using poke trajectories, similarly to \cite{Rimon2025ArtificialPalpation}, but with various orientations of the sensor, which are fixed throughout the motion. For data that better resembles human motion, we recorded several motion primitives using teleoperation, and tracked them using position control to generate \texttt{data-primitive}. Finally, test \texttt{data-handheld} was generated by manual motion, with poses recorded using a fiducial-based pose tracking device (see Appendices \ref{app:pose_estimation_system} and \ref{app:primitives} for details).
Following \cite{Rimon2025ArtificialPalpation}, we subsample the sensor measurements $1:50$ in \texttt{data-poke}.  For \texttt{data-primitive}, we subsampled the measurements $1:10$, which we found to work better due to
more complex motion. Each phantom in \texttt{data-poke} and \texttt{data-primitive} was poked using $100$ trajectories, while in \textit{data-handheld}, with 70. For all datasets, we concatenate the trajectories as input to \name{}.

\textbf{Automatic Data Collection:} To speed up data collection, we designed a motorized scissor lift to automatically reorient an insert within its shell as seen in \cref{fig:lift_down,fig:lift_up} (full details are in Appendix \ref{app:data_collection}). Using $3$ lifts and a Franka Panda manipulator, we collected data almost continuously for $\sim30$ days, resulting in $\sim800$ hours of tactile data, over 7$\times$ the size of \texttt{data-v1}.

\subsection{Tactile Imaging Training and Data}
\label{section:imaging_train_data}
The output of the \name\ image reconstruction model is either a $2$D or $3$D image. Similarly to \cite{Rimon2025ArtificialPalpation}, we train the model to predict images of the internal structure of the insert being palpated, obtained using MRI, as available in \texttt{data-v1} \citep{rimon2025datav1}. A straightforward loss function for image reconstruction is cross entropy between the predicted image and the MRI ground truth. However, especially for $3$D images, we found that our data suffers from severe class imbalance, as the inclusion is relatively small compared to the gel filling of the insert. We therefore opted for a combination of the Focal loss~\citep{lin2017focal} and Dice loss \citep{milletari2016v}, which we found to empirically improve performance, full details are in \refapp{app:class_imb}.

Ground truth MRI for the inserts we produced was obtained using a $3$T MRI (Siemens Prisma) system with a 64-channel coil, using the protocol described in \cite{Rimon2025ArtificialPalpation}. Different from \cite{Rimon2025ArtificialPalpation}, which only considered a $2$-dimensional horizontal slice of the scan, we consider the full $3$D volume. Full details on preparing the ground truth images are in \refapp{sec:ground_truth_processing_and_alignment}.

\subsection{Hand-held Tactile Imaging}
\label{section:method_handheld_imaging}
Considering applications of tactile imaging, the hand-held form factor is essential for clinical usage, as it allows the physician to maneuver and maintain patient contact during examination, similar to the workflow in ultrasound imaging. It may also be important for home or bedside devices. Next, we detail a proof-of-concept hand-held tactile imaging system.

\textbf{Pose Estimation and Generalization}
Our approach requires the sensor pose at each tactile measurement. In practice, sensor pose must be estimated, and we evaluated the robustness of our imaging to sensor pose noise. We found that when noise is i.i.d. Gaussian, our method remains stable to errors up to $\sim 1$ mm in position and $\sim0.1$ radian in orientation. Based on these results, we designed a fiducial-based pose estimation device with average pose error that is smaller than the stability thresholds above, as detailed in Appendix \ref{app:pose_estimation_system}. Using our system, we can palpate general objects manually to collect test data. Further, to generalize imaging to objects with different dimensions from the training objects, we extend the set of particles in the image decoder to include particles that cover a larger volume at test time. Since particles share weights, we can add an arbitrary number of particles, so long as we maintain a similar spatial resolution to the training data.

\textbf{Uncertainty Estimation Guidance}
Hand-held imaging devices such as ultrasound probes are known to be \textit{operator dependent}, a generally undesirable quality~\cite{european2020position}. To mitigate operator dependence, we propose to provide online feedback on imaging uncertainty, potentially guiding the operator to collect additional samples in uncertain regions. The \name\ architecture naturally accommodates \textit{local} uncertainty estimates: for each pixel in the image, we measure uncertainty using the Shannon entropy of the class probabilities. Since each pixel is affected only by local representations, the uncertainty is only affected by samples collected in each pixel's vicinity.

%% file: sections/experiments.tex
\section{Experiments} \label{section:experiments}

\begin{figure}[t]
    \centering
    \begin{subfigure}[c]{0.35\linewidth}
        \centering
        \includegraphics[width=\linewidth]{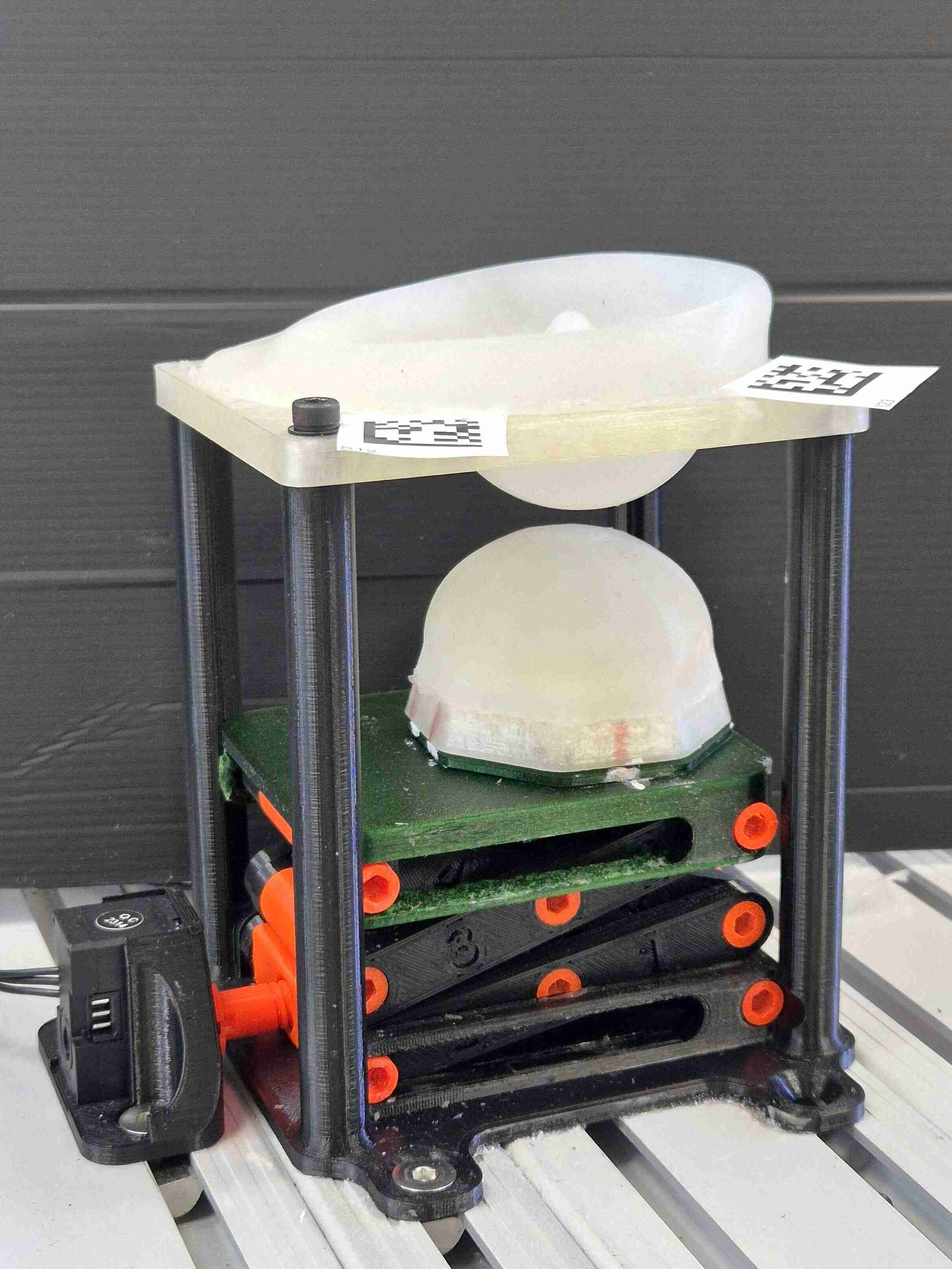} 
        \caption{}
        \label{fig:lift_down}
    \end{subfigure}
    \hfill
    \begin{subfigure}[c]{0.35\linewidth}
        \centering
        \includegraphics[width=\linewidth]{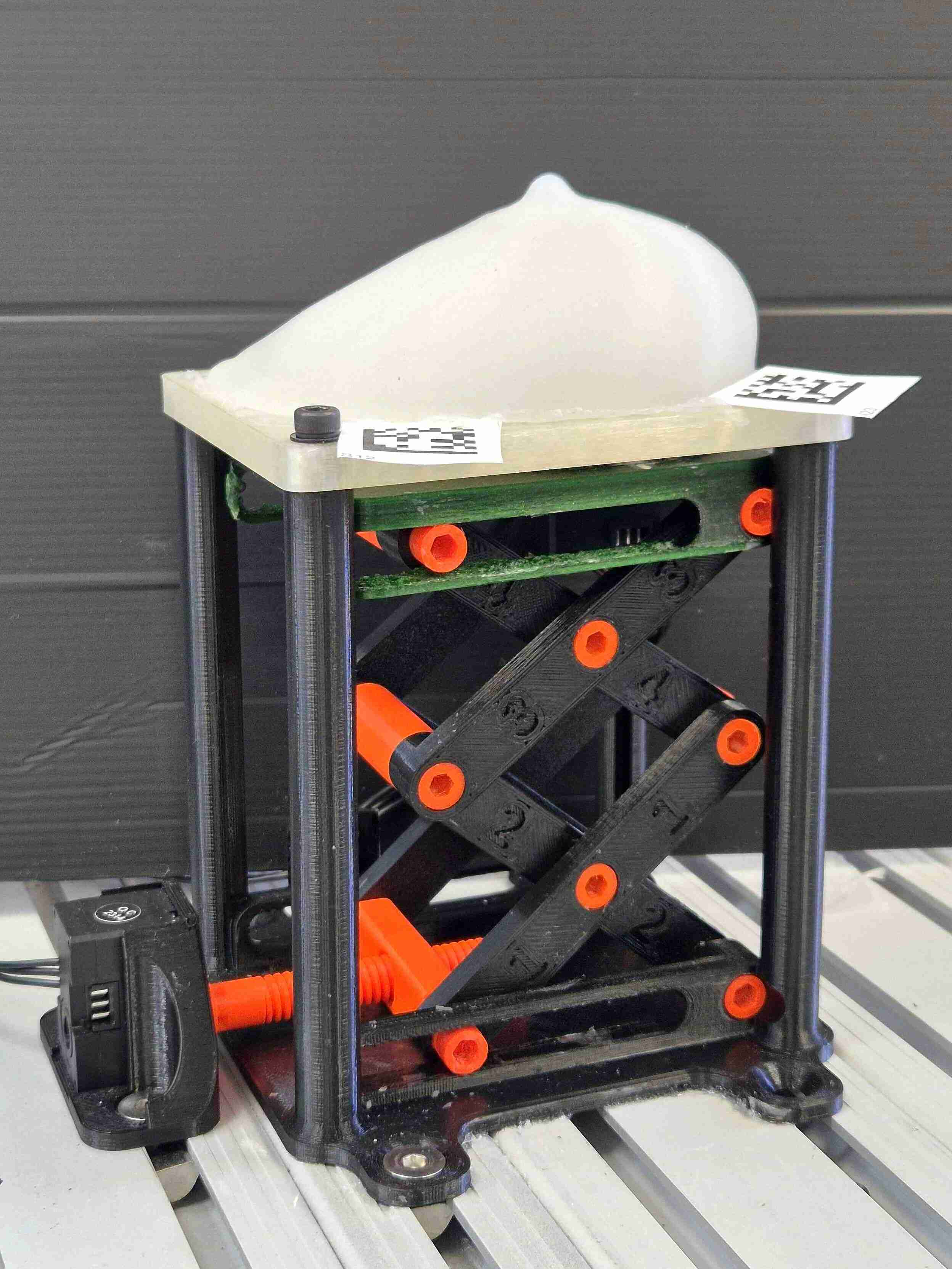} 
        \caption{}
        \label{fig:lift_up}
    \end{subfigure}
    \hfill
    \begin{subfigure}[c]{0.23\linewidth}
        \centering
        \includegraphics[width=\linewidth]{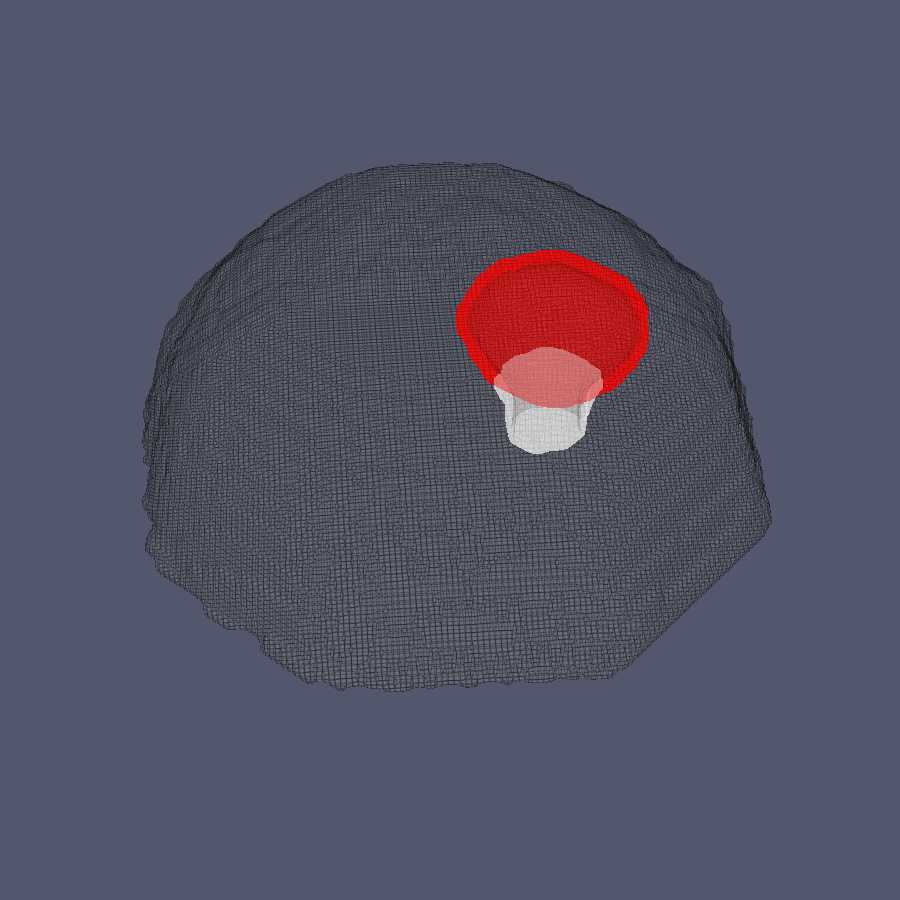}\\[0.07cm]
        \includegraphics[width=\linewidth]{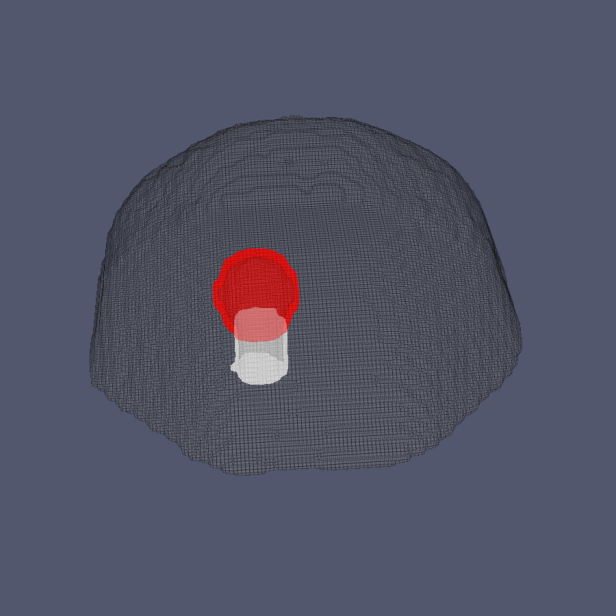}
        \caption{}
        \label{fig:3D_ground_truth_example}
    \end{subfigure}

    \caption{{\textit{(a)} and \textit{(b)} Insert being rotated and pushed into the shell in the different states of the scissor lift \textit{(c)} 3D ground truth for two inserts after post-processing of its MRI scan.
    \textit{Blue}: background.
    \textit{Gray}: insert.
    \textit{Red}: inclusion.
    \textit{White}: supporting pillar.}}
    \label{fig:phantom_and_insert}
\end{figure}

We conduct a series of experiments to study the following. \textit{(1)} Qualitative and quantitative assessment of 3D tactile imaging using our learning based approach and datasets. \textit{(2)} Can \name\ yield zero-shot compositional generalization? \ \textit{(3)} Is \name\ effective for local uncertainty estimation? and \textit{(4)} Can our system and data be used to demonstrate a proof-of-concept real-time hand-held tactile imaging?    

Throughout all experiments, we report the standard deviation by running 3 random seeds over train/test splits and random weights initialization.

\subsection{3D Tactile Imaging}
\label{sec:exp_2d_vs_3d}

As discussed in \ref{section:architecture}, we extend \citep{Rimon2025ArtificialPalpation} to 3D imaging allowing for richer visualization of the size, shape, and position of internal inclusions. 

An example of our 3D ground-truth is shown in \cref{fig:3D_ground_truth_example}. Our inserts (which are positioned inside breast-shaped shells) are filled with gel, with an outer skin made of soft silicone. In the insert, we place a round inclusion connected by a narrow pillar to the base of the insert, both made out of soft silicone. For visual clarity, we colour the pillar differently from the round inclusion, using an ad-hoc image processing method, described in \refapp{sec:ground_truth_processing_and_alignment}. The pillar is not touched by the sensor during poke motions; yet, it can be reconstructed, since the method was trained on ground truth images that include the pillars. This is a demonstration of the learned \textit{inference} capabilities of our approach. In the following, we quantitatively compare the accuracy of our new imaging results with the 2D imaging technique of \citep{Rimon2025ArtificialPalpation}.

\begin{table}[t]
\begin{small}
\centering
\caption{Comparison of global representation training on 2D and 3D labels. The first three rows are for 2D slice metrics and the rest are 3D metrics.}
\vspace{0.2cm}
\label{tab:2D_vs_3D}

\renewcommand{\arraystretch}{1.1}
\setlength{\tabcolsep}{3pt}

\begin{tabular}{p{0.4cm}lccc}
\toprule
 &  
 & \textbf{2D Training} 
 & \textbf{3D Training} 
 & \textbf{Average Pred.} \\
\midrule

\multirow{3}{=}{\centering\rotatebox{90}{\textbf{2D}}}
 & \textbf{$\Delta_{\text{Diameter}}$ [mm] $\downarrow$} & $2.7 \pm 0.1$ & $1.8\pm0.0$ & $2.02$ \\
 & \textbf{$\Delta_{\text{CoM}}$ [mm] $\downarrow$ }     & $2.1\pm 0.1$ & $1.7\pm0.1$ & $12.1$ \\
 & \textbf{F1 Score [\%] $\uparrow$  }                  & $79.9\pm 0.8$ & $85.2\pm0.8$ & -- \\
\midrule
\multirow{3}{=}{\centering\rotatebox{90}{\textbf{3D}}}
 & \textbf{$\Delta_{\text{Diameter}}$ [mm] $\downarrow$} & -- & $0.8\pm0.0$ & $1.44$ \\
 & \textbf{$\Delta_{\text{CoM}}$ [mm] $\downarrow$}      & -- & $1.7\pm0.1$ & $12.09$ \\
 & \textbf{F1 Score [\%] $\uparrow$ }                    & -- & $83.5\pm0.7$ & -- \\

\bottomrule
\end{tabular}
\end{small}
\end{table}

\begin{figure*}[t]
    \centering
    \includegraphics[width=.75\textwidth]{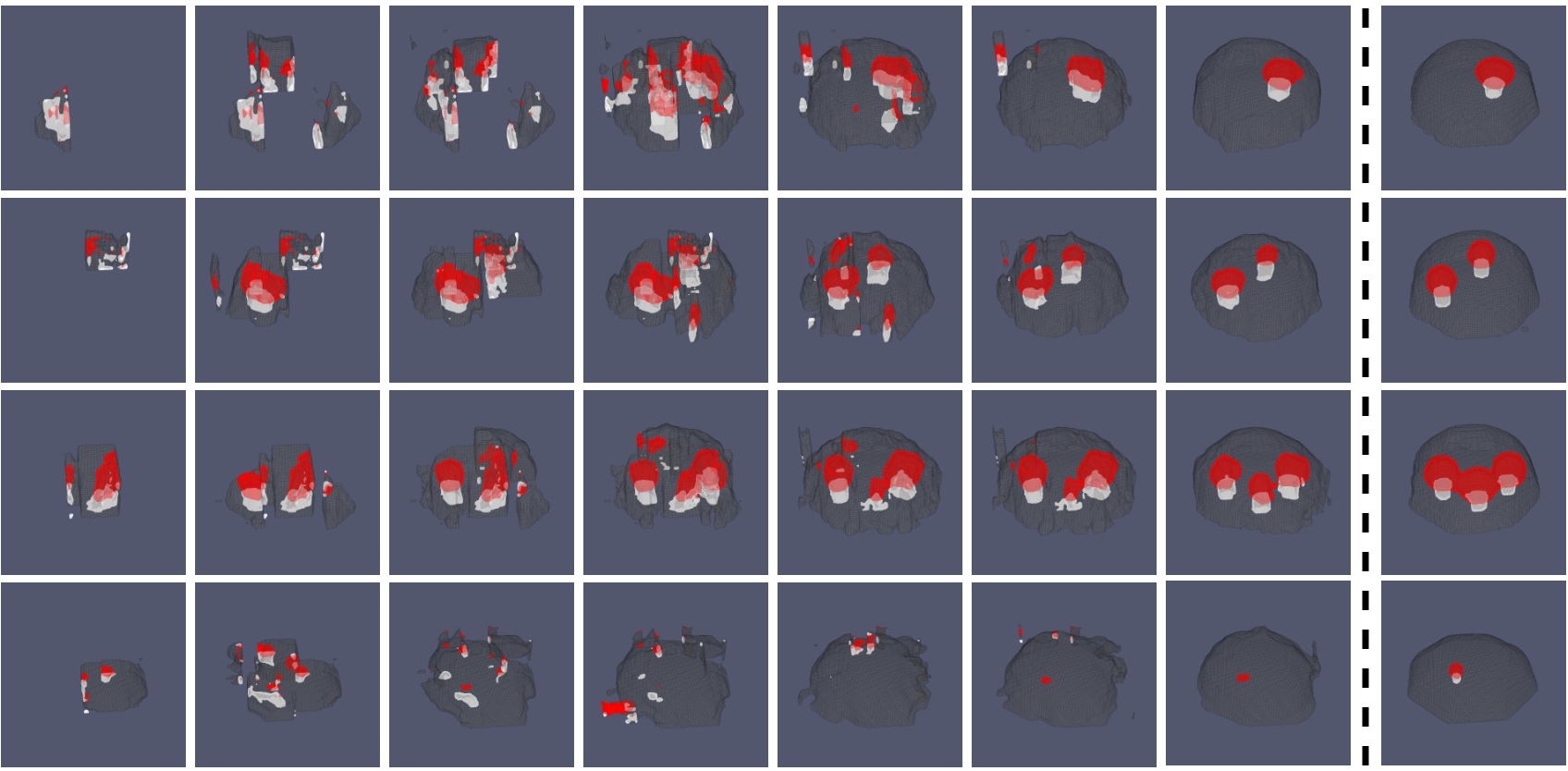}
    \caption{Prediction of \name\ for intermediate steps of four input sequences (steps were chosen to demonstrate development of the image over time). The ground-truth is provided on the right side. Different rows show: \textit{(1)} single inclusion, \textit{(2)} two inclusions, \textit{(3)} three connected inclusions, \textit{(4)} larger inclusion. As described in the text, the model was trained on single inclusions only.}
    \label{fig:less_results}
\end{figure*}

\begin{figure}[t]
    \centering
    \includegraphics[width=.85\linewidth]{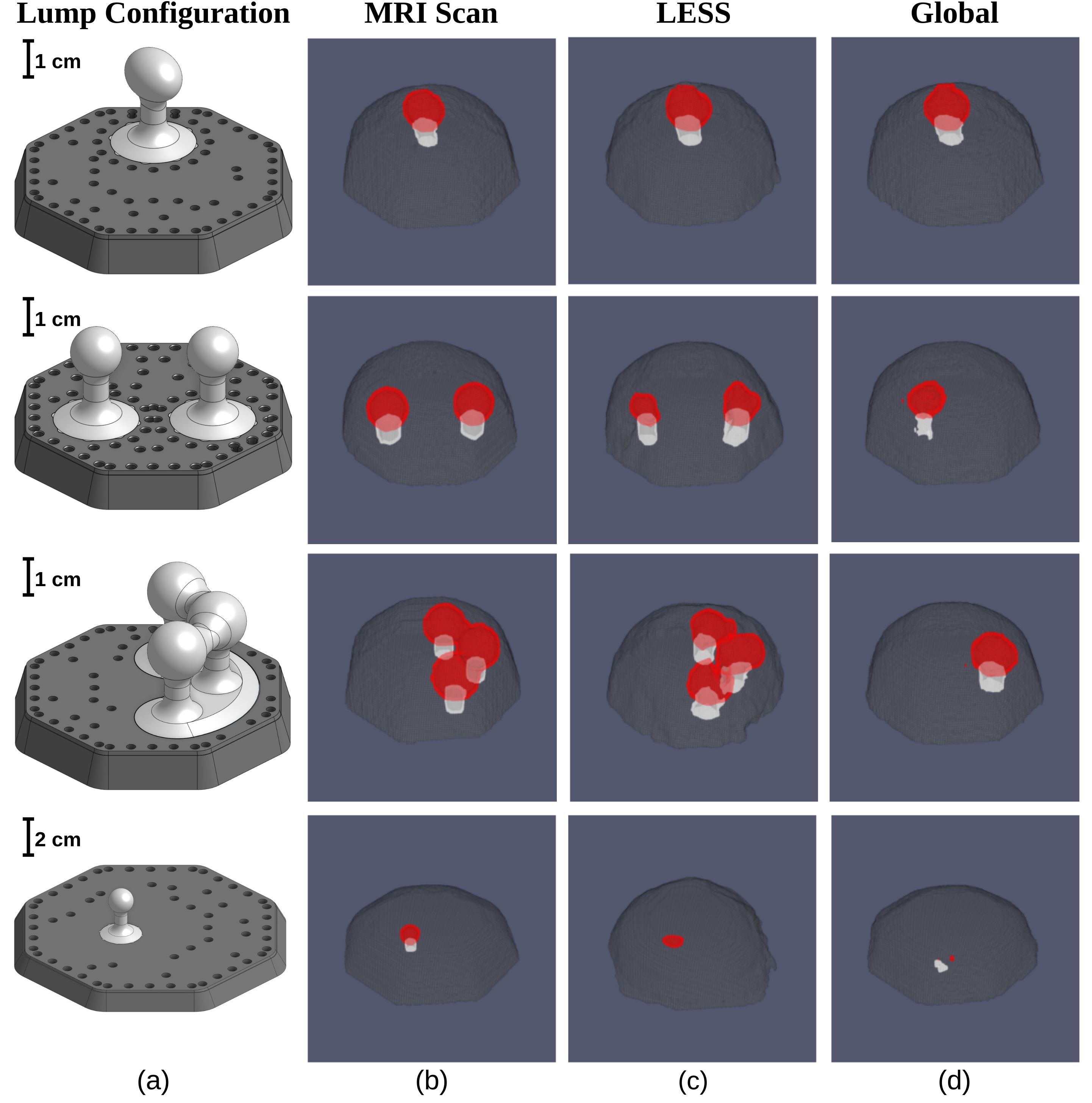}
    \caption{Comparison between multiple samples from our in-distribution (first row) and out-of-distribution (last three rows) inserts. See text for details. Columns show: \textit{(a)} 3D CAD \textit{(b)} 3D MRI ground-truth \textit{(c)} image reconstruction using \name\ \textit{(d)} image reconstruction using \texttt{GLOBAL}.}
    \label{fig:pred_comp}
\end{figure}

We begin by comparing our 3D reconstruction approach with the state-of-the-art approach of \citep{Rimon2025ArtificialPalpation}, and study how the dimensionality of the data affects reconstruction quality. To separate the contribution of the data dimensionality from other architectural improvements in \name, we trained the representation of \citep{Rimon2025ArtificialPalpation} on \texttt{data-poke} with 2D and 3D MRI labels, where the 3D training is based on our method in \cref{section:imaging_train_data}.

For the 2D labels, we report the F1 score, inclusion Center-of-Mass (CoM) error, and inclusion diameter error. For the 3D labels we report these metrics on a constant height slice and the 3D version of these metrics. Exact definition of the metrics can be found in \refapp{sec:app_metrics}.
To connect our results to easily interpretable quantities, we also report the metrics for predicting the average diameter and CoM over the dataset.

Surprisingly, \cref{tab:2D_vs_3D} shows that incorporating 3D predictions leads to consistent improvements in 2D metrics compared to the results reported in \citep{Rimon2025ArtificialPalpation}. We hypothesize that this improvement originates from the additional geometric prior provided by the 3D representation. When sufficient training data is available, learning from volumetric information encourages the model to capture spatial regularities that are not observable in isolated 2D slices, such as consistency across adjacent slices. Thus, in addition to producing more interpretable visualizations, 3D data also yields more accurate reconstructions.

For the rest of the experiments, we train all models with the 3D labels, but report the 2D metrics, which are easier to interpret, and we term the global baseline of \citep{Rimon2025ArtificialPalpation}, trained on the 3D labels of \texttt{data-poke} as \texttt{GLOBAL}.

\begin{figure*}[t]
    \centering    \includegraphics[width=.75\textwidth]{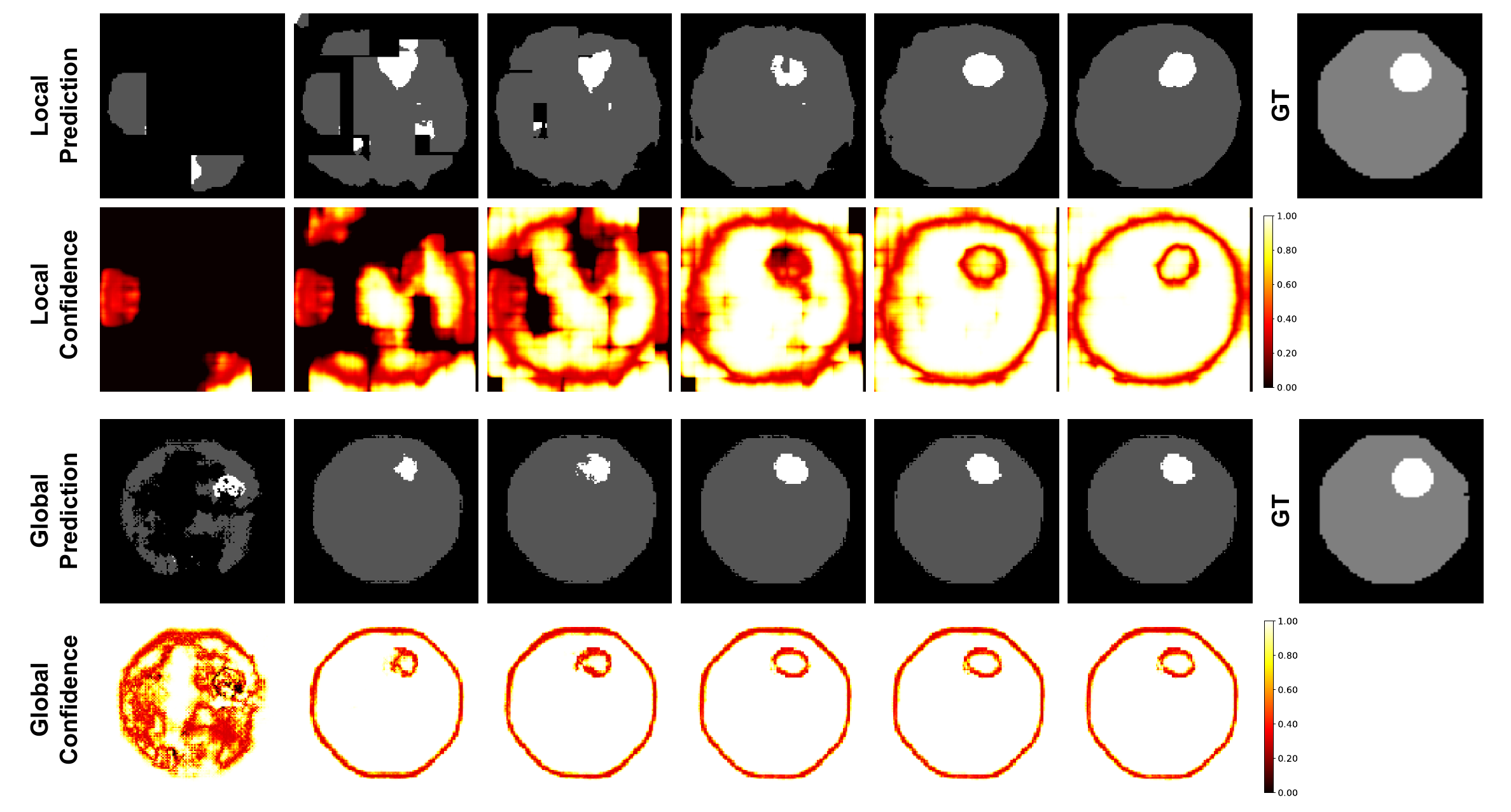}
    \caption{Prediction and confidence of \name\ compared to \texttt{GLOBAL} for intermediate steps of the same input sequence.
   }
    \label{fig:confidence}
\end{figure*}

\subsection{Zero-Shot Compositional Generalization with \name}

One of the core benefits of \name\ is the emergent compositional generalization (as discussed in \cref{section:architecture}).
To demonstrate and evaluate this, we manufactured a set of specialized phantoms, unseen during training, that test compositional generalization to changes in the shape of the object and the number of inclusions inside it. Accordingly, we split the specialized phantoms into two categories - \texttt{multiple} and \texttt{large}.
The \texttt{multiple} set corresponds to phantoms with multiple inclusions in a single insert, and the \texttt{large} set corresponds to a phantom with the same height, but a base with a $4\times$ larger area than the regular phantoms, and with single inclusions with similar shapes and sizes to the regular phantoms. Full details of the specialized phantoms can be found in \refapp{sec:app_manufacturing_details}. We emphasize that in this section, we trained models on the \texttt{data-poke} dataset, which only includes phantoms with \textit{single} inclusions. Thus, phantoms in \texttt{multiple} are essentially out of distribution.
We collected test interactions on the specialized phantoms with the same controller as in \texttt{data-poke}.

\begin{table}[h]
\centering
\caption{Results of \texttt{GLOBAL} and \name\ on in-distribution single inclusion and out-of-distribution multiple inclusions and large phantom. In the \texttt{multiple} set, we report the total area error instead of the diameter error. Due to their different inclusion composition, the metrics of different insert sets are not comparable. We clarify that in the \texttt{large} set, the diameter and CoM errors are expected to be larger.}
\vspace{0.2cm}
\label{tab:less_vs_global}

\renewcommand{\arraystretch}{1.1}

\begin{tabular*}{0.8\linewidth}{p{0.4cm} @{\hspace{5pt}} l @{\extracolsep{\fill}} c c}
\toprule
 &
 & \textbf{GLOBAL}
 & \textbf{\name\ (ours)} \\
\midrule

\multirow{3}{*}{\rotatebox{90}{\textbf{\parbox{1cm}{\centering Single}}}}
 & \textbf{$\Delta_{\text{Diameter}}$ [mm] $\downarrow$} & $1.8 \pm 0.0$ & $1.5 \pm 0.1$ \\
 & \textbf{$\Delta_{\text{CoM}}$ [mm] $\downarrow$ }       & $1.7 \pm 0.1$ & $1.7 \pm 0.1$ \\
 & \textbf{F1 Score [\%] $\uparrow$                    }    & $85.2 \pm 0.8$ & $83.5 \pm 0.6$ \\
\midrule

\multirow{3}{*}{\rotatebox{90}{\textbf{\parbox{1cm}{\centering Multiple}}}}
 & \textbf{$\Delta_{\text{Area}}$ [mm$^2$] $\downarrow$} & $78.4 \pm 5.1$ & $46.1 \pm 2.9$ \\
 & \textbf{$\Delta_{\text{CoM}}$ [mm] $\downarrow$}        & $7.4 \pm 0.4$ & $4.0 \pm 0.0$ \\
 & \textbf{F1 Score [\%] $\uparrow$ }                     & $44.6 \pm 1.3$ & $71.1 \pm 0.0$ \\
\midrule

\multirow{3}{*}{\rotatebox{90}{\textbf{\parbox{1cm}{\centering Large}}}}
 & \textbf{$\Delta_{\text{Diameter}}$ [mm] $\downarrow$} & $9.0 \pm 1.8$ & $2.7 \pm 1.1$ \\
 & \textbf{$\Delta_{\text{CoM}}$ [mm] $\downarrow$}        & $26.8 \pm 2.8$ & $2.3 \pm 0.2$ \\
 & \textbf{F1 Score [\%] $\uparrow$ }                     & $1.1 \pm 0.5$ & $71.3 \pm 3.6$ \\

\bottomrule
\end{tabular*}
\end{table}

We trained \name\ on \texttt{data-poke} similarly to \texttt{GLOBAL}. \name\ is able to effectively reconstruct complex internal structure as highlighted in \cref{fig:less_results}.
We compare \name\ and \texttt{GLOBAL} on other types of phantoms at test time. As seen in \cref{tab:less_vs_global}, \name\ performs slightly worse compared to \texttt{GLOBAL} on in-distribution single inclusions, since each image patch that \name\ generates is affected by less information (due to the receptive fields), making the prediction problem more difficult. However, \name\ significantly outperforms the baseline on the out-of-distribution benchmark. 

As shown in \cref{fig:pred_comp}, \texttt{GLOBAL} does not generalize to the \texttt{multiple} set, and generates an image with only one of the inclusions. This is expected, as training on single inclusion data is not sufficient for generalization to more complex inner structure when using global representations. On the other hand, \name\ is able to generalize the local patterns of single inclusions in the data to accurately generate multiple inclusions. The \texttt{large} set is completely out of distribution for \texttt{GLOBAL} (as it was never trained on poses that are encountered in this set). On the other hand, due to \name's centering process, it is able to generalize to this set as well.   
We believe these results show the potential of \name\ for tactile imaging of generally structured objects.  

\begin{figure}[h]
  \centering
  \includegraphics[trim={3cm 5cm 15cm 15cm},clip,width=0.99\linewidth]{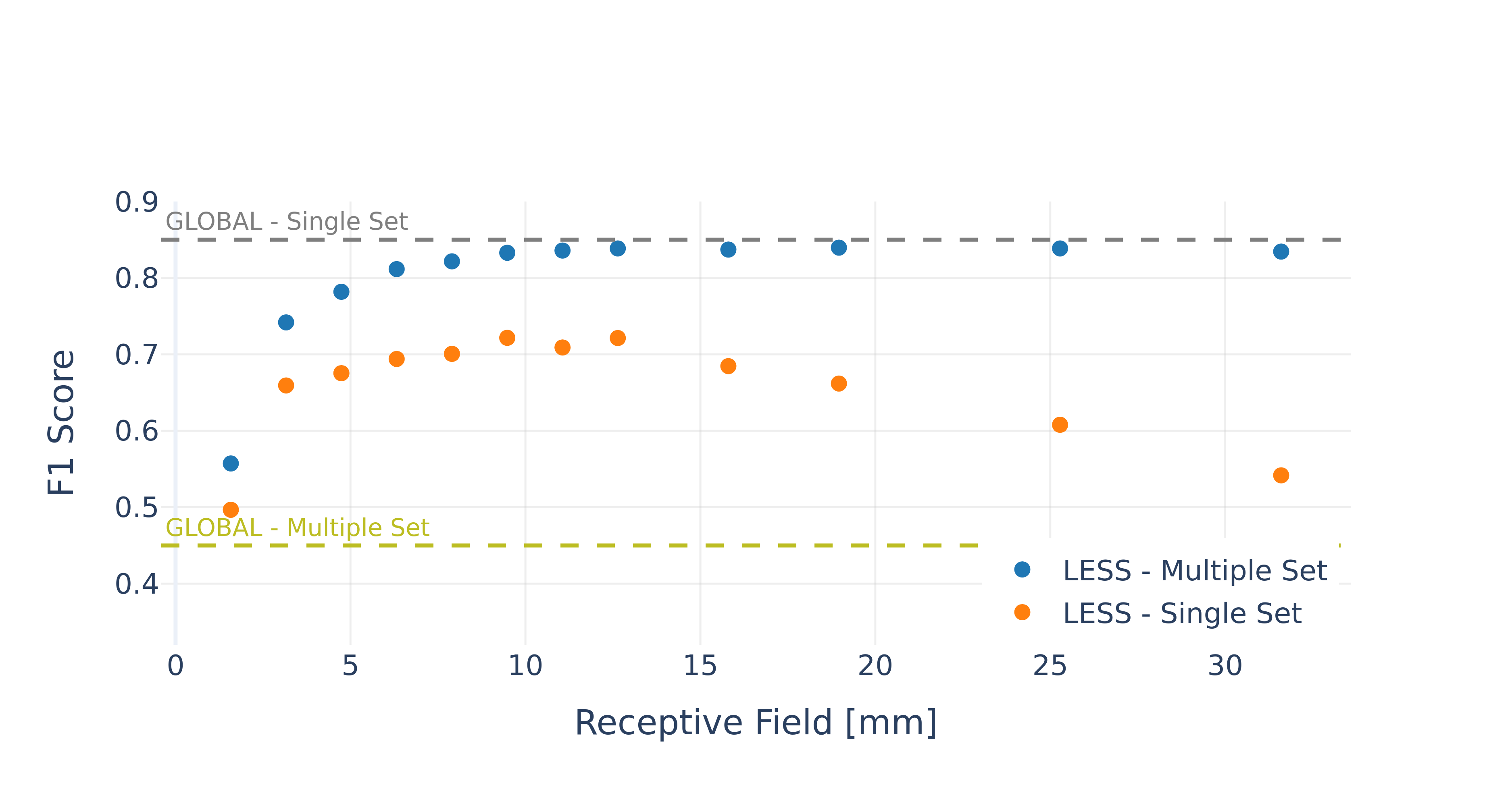}
  \caption{\name\ performance for different receptive field sizes.} \label{fig:ablation_receptive_field}
\end{figure}

\begin{figure}[t]
    \centering
    \includegraphics[trim=0 7cm 0 7cm, clip, width=.99\linewidth]{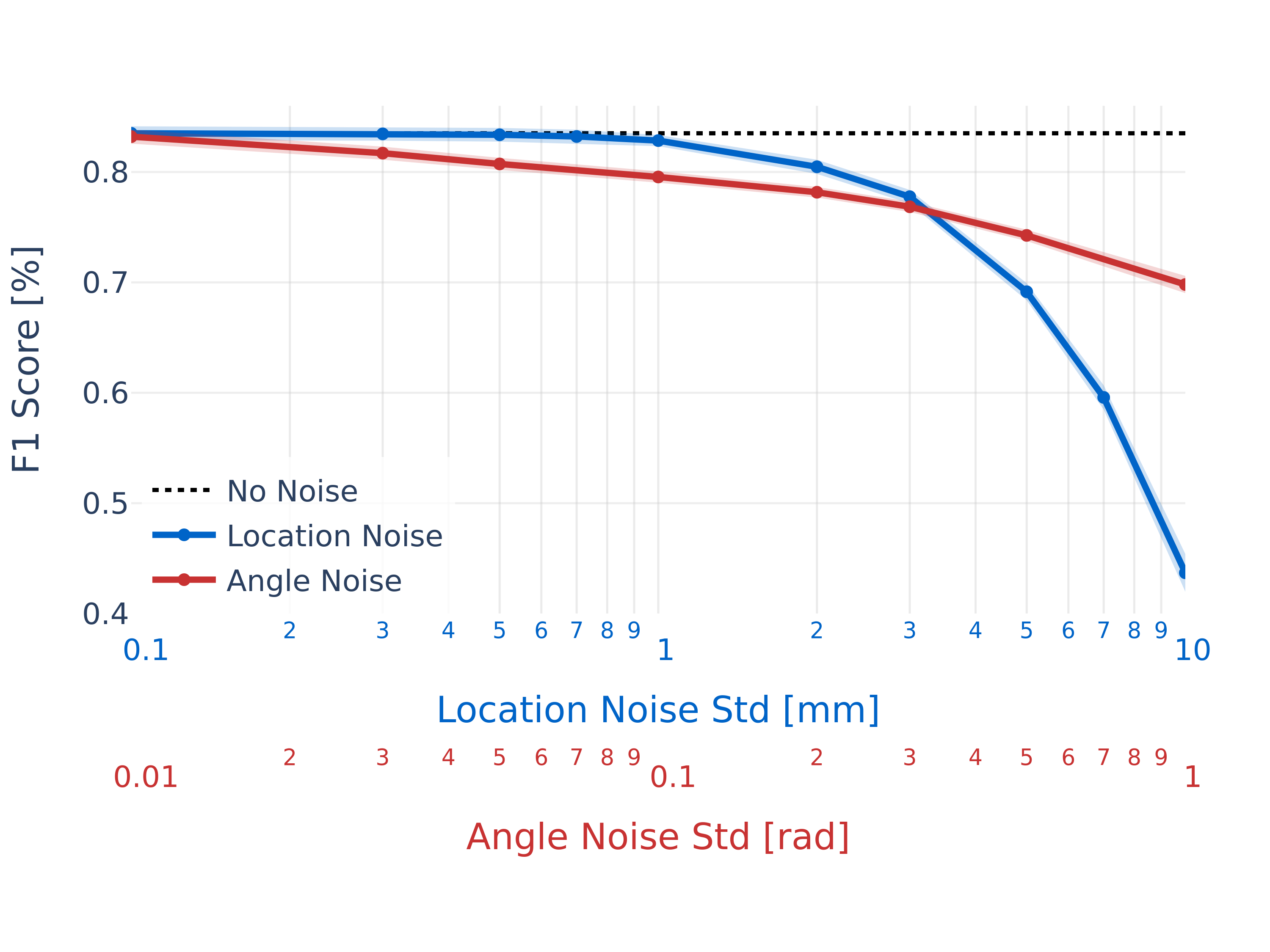}
    \caption{Effect of pose noise on the F1 score of \name{}.
   }
   \vspace{-0.5cm}
    \label{fig:noise_robustness}
\end{figure}

\begin{figure*}[t]
    \centering
    \includegraphics[width=.99\textwidth]{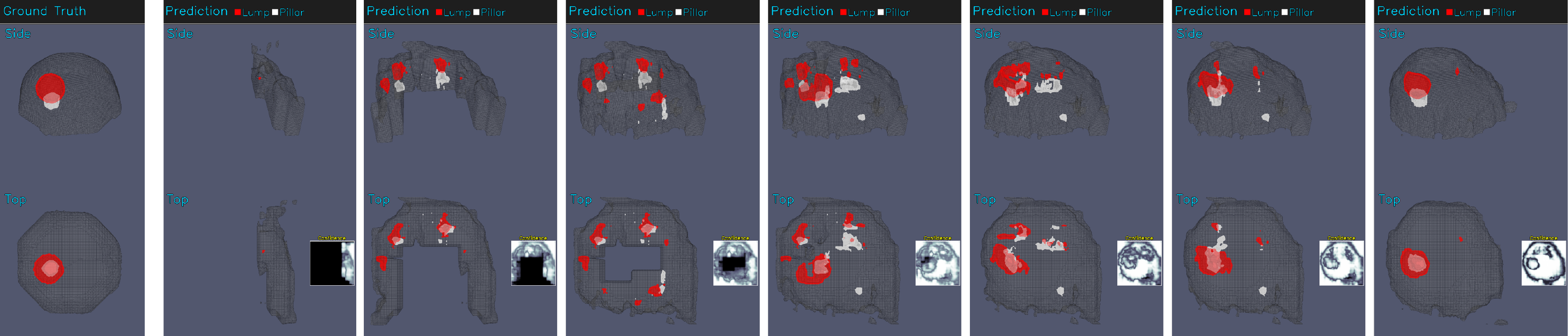}
    \caption{The monitor presented to the operator at different steps while using our hand-held system. On the far left, we show the ground truth. The operator can observe the prediction and the uncertainty estimates (bottom right box in each plot). 
   }
   \vspace{-0.2cm}
    \label{fig:hand_held}
\end{figure*}

\subsection{Receptive Field Tuning}

An important hyperparameter of \name\ is the receptive field size $\recfield$. As can be seen in \cref{fig:ablation_receptive_field}, increasing the receptive field allows more information to be encoded in each particle, improving the performance on the \texttt{single} set and eventually achieving similar accuracy to \texttt{GLOBAL}. The tradeoff in increasing $\recfield$ is reducing the local invariance, thereby reducing compositional generalization, as evident by the degraded performance on the \texttt{multiple} set in \cref{fig:ablation_receptive_field}. Finally, we observe that $\beta\approx10[mm]$ results in the best performance on the \texttt{multiple} set, without compromising accuracy on the \texttt{single} set; we used this $\recfield$ value throughout our experiments.

\subsection{Local Uncertainty Estimation}
We qualitatively evaluate uncertainty estimation using \name\ and compare with \texttt{GLOBAL}. For both methods we use the Shannon entropy for pixel level uncertainty, as described in \cref{section:method_handheld_imaging}; the only difference is in the neural network that generates pixel probabilities.
In \cref{fig:confidence} we show the image reconstruction and uncertainty of both models during intermediate steps of reconstructing an image from the same tactile input sequence. As can be seen, \name\ produces a significantly more informative uncertainty estimation, showing low confidence on unobserved areas, while \texttt{GLOBAL} collapses to a very high confidence early on, before many areas of the phantom have been touched. 
To further understand these results, note that \texttt{GLOBAL} yields high confidence in areas where it lacks sufficient data, such as those outside the insert area. The reason is that these pixels are always outside the insert area in the training data. In comparison, \name\ produces very low confidence in these coordinates, until it receives sufficient data in their vicinity.

\subsection{Hand-Held Tactile Imaging} 
Finally, we investigate whether our tactile imaging system can be used as a hand-held device that produces accurate imaging in real-time. The two main challenges in this task are tracking the pose of the tactile sensor accurately, and the distribution shift between sensor motion in the robot-operated data collection and manual hand-held motion. In this section we first seek to test the robustness of \name\ to inaccurate pose, which is crucial for certain design choices in our pose estimation system. Then, we show that by training on \texttt{data-primitive} we substantially reduce the distribution shift that occurs when testing on human motions.

\textbf{Robustness to Inaccurate Pose} To determine the required system accuracy, we evaluated the sensitivity of the model to errors in the sensor pose. Two types of noise were injected: translational and angular. $\spose_\ttime \in \mathbb{R}^6$ is the true pose, where the first $3$ components are $x-y-z$ positions, and the last $3$ components are pitch-roll-yaw angles. We modeled the noise as additive i.i.d. zero-mean Gaussian noise: $\estpose_\ttime = \spose_\ttime + \mathcal{N}(0,\textrm{diag}(\bar{\sigma}))$, where $\bar{\sigma} = [\sigma_{pos}, \sigma_{pos}, \sigma_{pos}, \sigma_{angle}, \sigma_{angle}, \sigma_{angle}]$. We evaluated tactile imaging results with varying noise in the test input poses, shown in \cref{fig:noise_robustness}. 
Based on these results, we have designed our pose estimation system as discussed in \cref{section:method_handheld_imaging}. We verify the noise level of our pose to be less than $1 mm$ and $0.02 rad$ and further measure end-to-end performance degradation due to pose error in \refapp{app:pose_estimation_accuracy}.

\textbf{Training on More Diverse Data} To reduce the distribution shift when testing on human motions in the hand-held setup, we propose to train \name\ on \texttt{data-primitive}. This dataset contains a much more diverse set of motions compared to the constant angle poke trajectories in \texttt{data-poke}. We found that combining \texttt{data-poke} with \texttt{data-primitive} as the training data yielded the best results. An ablation study on the training data is presented in \cref{tab:primitives_training}.

\begin{table}[h]
\footnotesize 
\centering
\caption{Training \name\ on \texttt{data-primitive}. We trained  on only \texttt{data-poke} (\textbf{Poke}), only \texttt{data-primitive} (\textbf{PRI}) and the union of \texttt{data-poke}, and \texttt{data-primitive} (\textbf{PRI + Poke}). All methods were tested on \texttt{data-primitive} only.}
\label{tab:primitives_training}

\setlength{\tabcolsep}{2pt} 

\begin{tabular}{p{2.3cm}ccc}
\toprule
\textbf{Method} & \textbf{$\Delta_{\text{Diameter}}$ [mm] $\downarrow$} & \textbf{$\Delta_{\text{CoM}}$ [mm] $\downarrow$} & \textbf{F1 Score [\%] $\uparrow$} \\
\midrule
\textbf{Poke}             & $6.3 \pm 0.6$ & $6.0 \pm 0.5$ & $45.8 \pm 0.6$ \\
\textbf{Pri}             & $2.8 \pm 0.3$ & $2.3 \pm 0.1$ & $74.7 \pm 1.2$ \\
\textbf{Pri + Poke}      & $2.0 \pm 0.2$ & $1.6 \pm 0.1$   & $80.9 \pm 0.8$ \\
\bottomrule
\end{tabular}
\end{table}

\textbf{Hand-Held Data} We finally tested \name\ on \texttt{data-handheld}. To improve the results visually, we performed a calibration of the threshold for classifying a pixel in the image as an inclusion or a supporting pillar, choosing a threshold of 0.8. We tested \name\ trained with \texttt{data-primitive} and \name\ trained without \texttt{data-primitive}. As can be seen in \cref{tab:handheld}, training on the combination of \texttt{data-primitive} and \texttt{data-poke}, helped mitigate the distribution shift when considering hand-held motion and resulted in improved performance. 
We achieve a $\Delta_{CoM}$ of $8.4[mm]$ compared to  $1.6[mm]$ in the robotic setup, about half a centimeter worse. Although our results show that tactile imaging using the robot-held sensor outperforms the hand-held setup, we believe that the results also show a positive trend: with more diverse robotic data, we expect to improve hand-held tactile imaging in the future.

\begin{table}[ht]
\footnotesize  
\centering
\caption{Testing \name\ on \texttt{data-handheld}.}
\label{tab:handheld}

\setlength{\tabcolsep}{2pt} 

\begin{tabular}{p{2.2cm}ccc}
\toprule
\textbf{Method} & \textbf{$\Delta_{\text{Diameter}}$ [mm] $\downarrow$} & \textbf{$\Delta_{\text{CoM}}$ [mm] $\downarrow$} & \textbf{F1 Score [\%] $\uparrow$} \\
\midrule
\textbf{w/ Pri, }  & $2.8\pm0.8$ & $8.4\pm0.7$ & $41.3\pm2.9$ \\
\textbf{w/o Pri, } & $5.2\pm0.5$ & $9.5\pm0.8$ & $34.7\pm 4.2$ \\
\bottomrule
\end{tabular}
\end{table}

\textbf{Real-Time Hand-Held Tactile Imaging} We provide a visualization of our proposed hand-held tactile imaging system operation in \cref{fig:hand_held}. A full demonstration video can be found on the project website (\href{ https://zoharri.github.io/LESS}{ \textit{zoharri.github.io/LESS}}).

%% file: sections/conclusion.tex
\section{Discussion} \label{section:conclusion}
\label{sec:conclusion}

Understanding artificial touch is a long-standing challenge in AI, with tactile imaging being an important application. In this work we take a step in this direction by proposing a neural network architecture tailored for spatio-temporal tactile data and new data collection techniques for making sense of soft objects. Technically, our work paves the way for learning-based tactile imaging that is hand-held, operator-independent, and applicable to arbitrary-shaped objects with composite internal structure. These capabilities have direct clinical value: operator independence is important for home-use devices, while detecting multiple inclusions can inform a physician’s decision to perform multiple biopsies, potentially improving patient outcomes.

A pressing question is clinical relevance of a method trained on synthetic data. In principle, with a large and diverse synthetic object dataset, learning-based tactile imaging may generalize to human tissue with sufficient accuracy. Future work must evaluate this hypothesis on human-subject data; our technical contribution in this work, in particular, the hand-held system we developed, will be important for clinical evaluation.

Beyond tactile imaging, our neural architecture may be useful for other robotic domains that involve interpreting complex spatio-temporal signals, such as learning object affordances from video~\citep{nagarajan2020ego}, or soft object manipulation~\citep{lin2021softgym}.

%% file: sections/acknowledgments.tex
\section*{Acknowledgments} \label{section:ack}
This work received funding from the European Union (ERC, Bayes-RL, Project Number
101041250). Views and opinions expressed are however those of the authors only and do not necessarily reflect those of the European Union or the European Research Council Executive Agency.
Neither the European Union nor the granting authority can be held responsible for them.

%% file: sections/appendix/appendix.tex
\clearpage
\appendix
\etocsettocstyle{\subsection*{}}{}
\localtableofcontents 

\input{sections/appendix/fabrication}

\input{sections/appendix/3d_gt}

\input{sections/appendix/3d_metrics}

\input{sections/appendix/focal_loss}

\input{sections/appendix/ssl_loss}

\input{sections/appendix/pose_estimation_details}

\input{sections/appendix/pose_estimation_accuracy}

\input{sections/appendix/general_data_collection}

\input{sections/appendix/primitives}

\input{sections/appendix/handheld_technical_details}

%% file: sections/appendix/fabrication.tex
\subsection{Phantom Fabrication Technical Details}
\label{sec:app_manufacturing_details}
\label{appendix:fabrication}
In \cite{Rimon2025ArtificialPalpation}, silicone was spread on the mold in much the same fashion as in the manufacturing of hollow chocolates. We chose to use a different process where a consistent, uniform thickness for the skin of the phantom is achieved by using two molds. We will describe the differences in the manufacture of each part in the following sections.

\begin{figure}[htbp]
     \centering
     \begin{subfigure}[b]{0.2\textwidth}
         \centering
         \includegraphics[width=\textwidth]{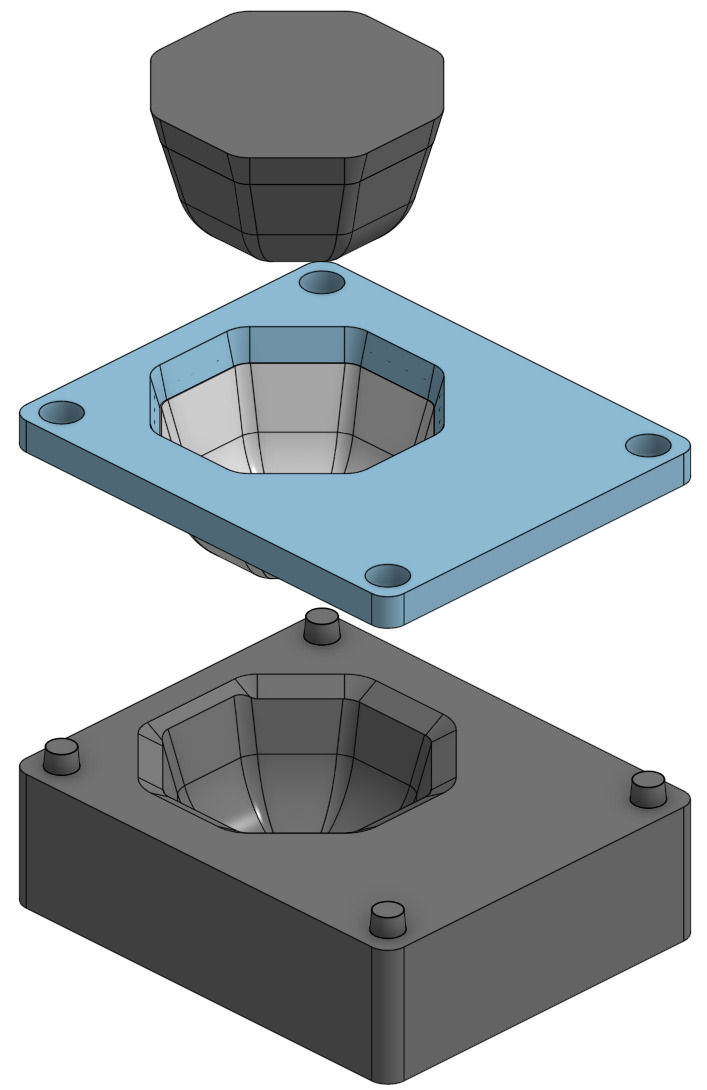}
         \caption{Cavity mold, exploded view with cured silicone attached to base}
         \label{fig:mould_shell_cavity}
     \end{subfigure}
     \hfill
     \begin{subfigure}[b]{0.2\textwidth}
         \centering
         \includegraphics[width=\textwidth]{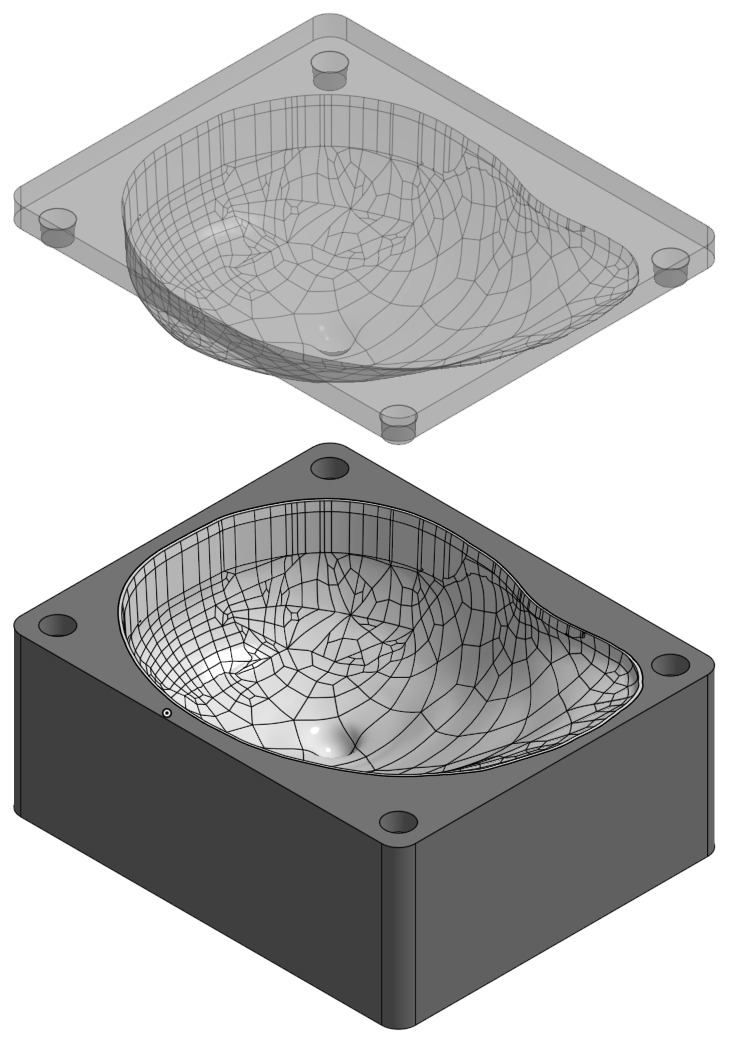}
         \caption{Outer skin mold}
         \label{fig:shell_mould_top}
     \end{subfigure}
     
     \vspace{1em} 

     \begin{subfigure}[b]{0.25\textwidth}
         \centering
         \includegraphics[width=\textwidth]{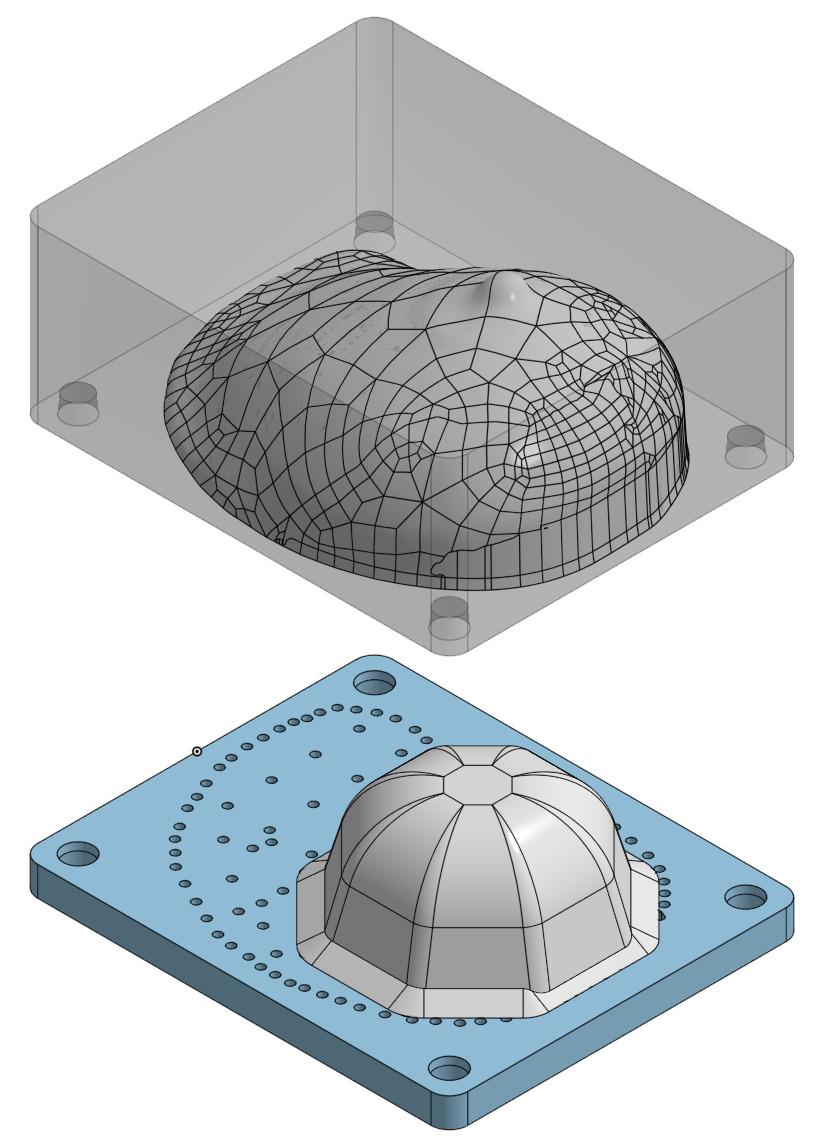}
         \caption{Shell mold final step}
         \label{fig:shell_exploded}
     \end{subfigure}

     \caption{Manufacturing process for the shell: We cast the cavity (a) and outer skin (b) in parallel (cured silicone in light grey). After curing, we pour silicone on the base plate and join it with the outer skin mold with cured silicone (c).}
\end{figure}

\textbf{Shell} In the \cite{Rimon2025ArtificialPalpation} process, silicone was added to a mold and put into a tumbler for an even coat. In our process we have two separate molds, one for the base with a cavity for the insert (see \cref{fig:mould_shell_cavity}), another for the top skin (see \cref{fig:shell_mould_top}). Note that the cavity is molded into the base plate. Silicone is poured into each of these molds in parallel. After the silicone has set, we do the following with each of the molds: For the top mold we remove only the top part while the silicone is still attached to the bottom part. For the cavity mold, we remove the silicone cavity while leaving it attached to the base plate as in \cref{fig:mould_shell_cavity}. We then pour silicone onto the base plate to achieve a thickness of a few millimeters and cover with the bottom part still attached to the mold, as in \cref{fig:shell_exploded} 

\begin{figure}[ht] 
    \centering
    \includegraphics[width=0.25\textwidth]{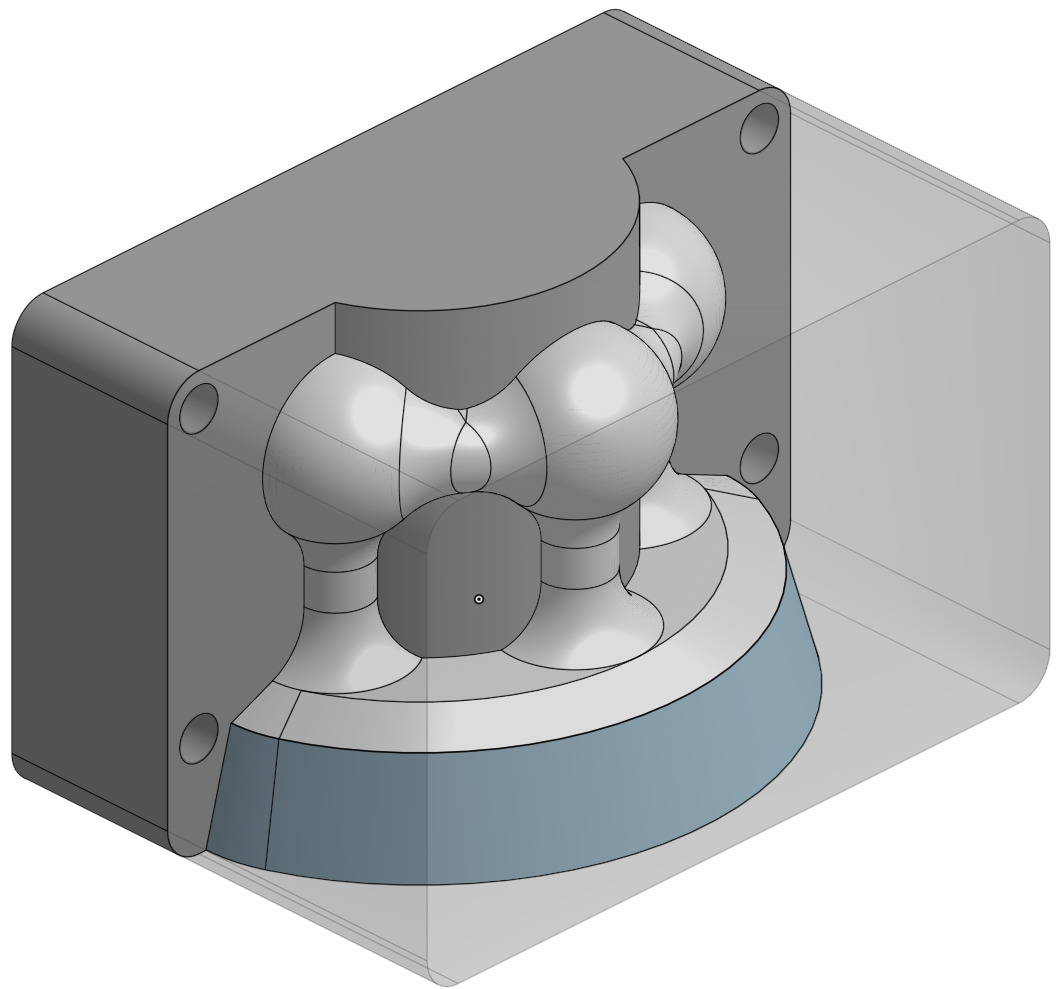}
    \caption{Two-part mold for three inclusions in a curved formation}
    \label{fig:mould_lump_curved}
\end{figure}

\begin{figure}[htbp]
     \centering
     \begin{subfigure}[b]{0.2\textwidth}
         \centering
         \includegraphics[width=\textwidth]{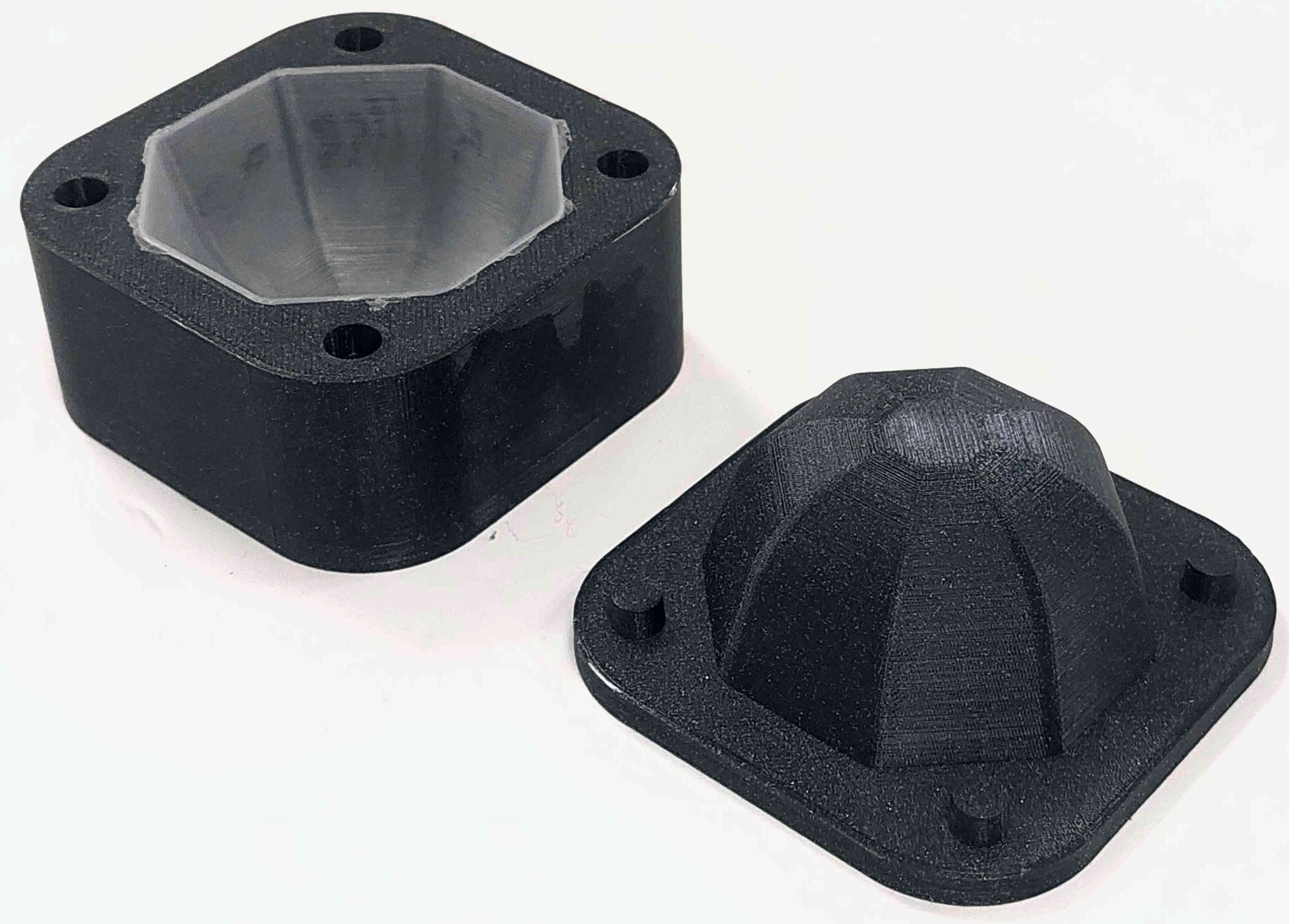}
         \caption{Molds for top skin and skin.}
         \label{fig:mould_real_insert_top}
     \end{subfigure}
     \hfill
     \begin{subfigure}[b]{0.2\textwidth}
         \centering
         \includegraphics[width=\textwidth]{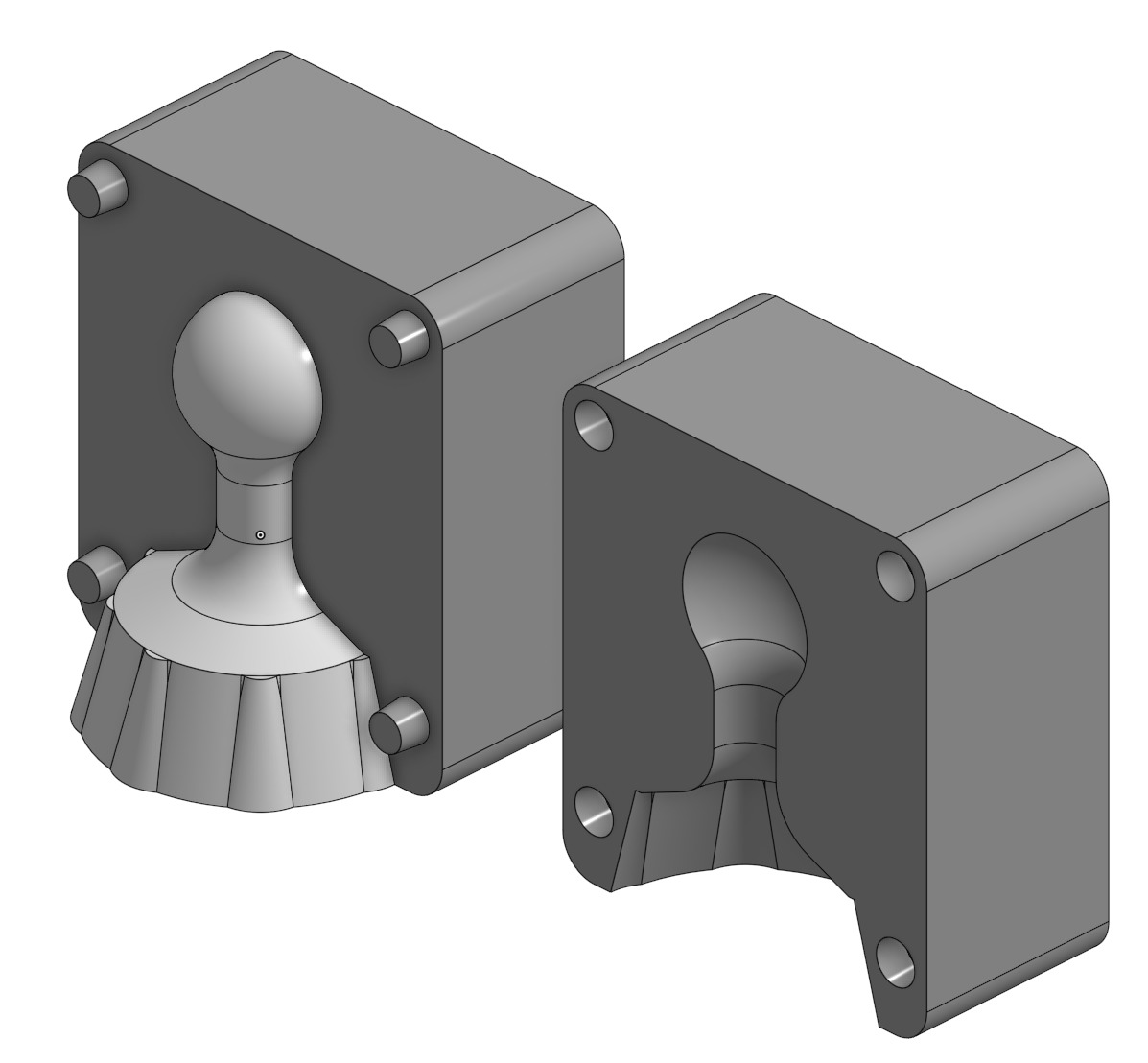}
         \caption{Mold for inclusion and ready inclusion}
         \label{fig:mould_lump_round}
     \end{subfigure}
     
     \vspace{1em} 

     \begin{subfigure}[b]{0.25\textwidth}
         \centering
         \includegraphics[width=\textwidth]{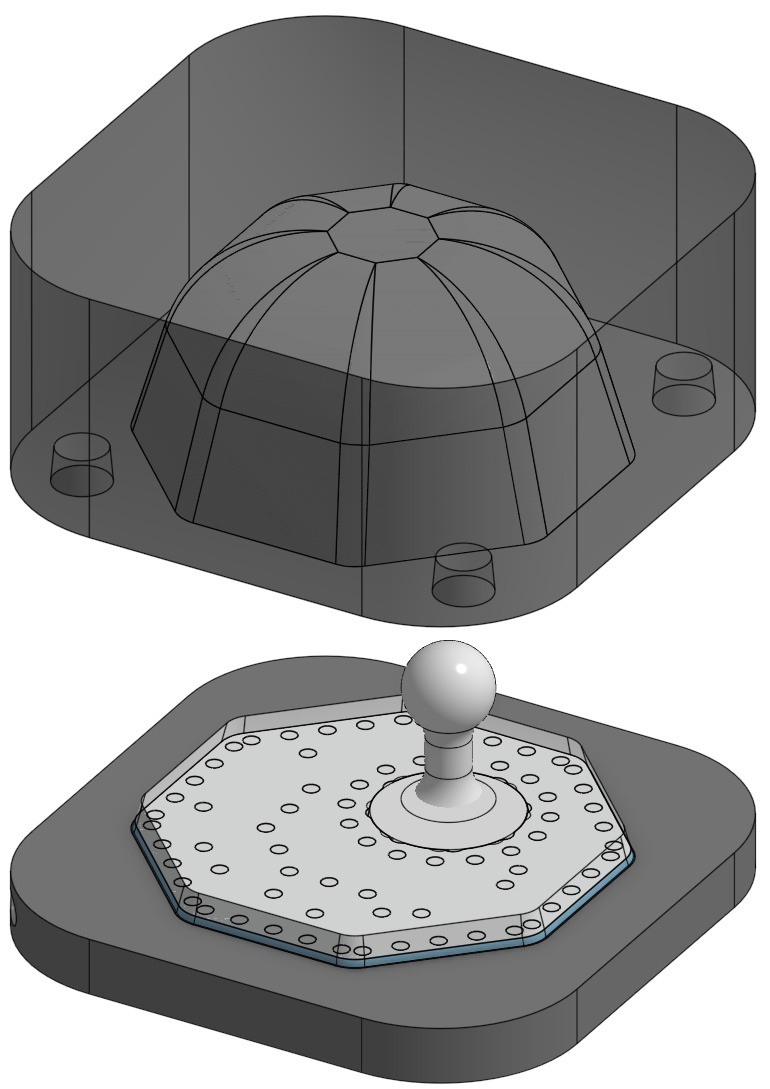}
         \caption{Final Step}
         \label{fig:insert_final_step}
     \end{subfigure}

     \caption{Steps of the insert manufacturing process. We create the top skin and insert in parallel (a and b). The final step of the manufacturing process insert is shown in (c). The top part has the silicone skin inside it. For the bottom part, we put in the insert, add a flange (in dark grey) and pour silicone (in transparent light grey).}
\end{figure}

\textbf{Insert}
As with the shell, we have separate molds, in this case one for the top part and another for the inclusion (see \cref{fig:mould_real_insert_top,fig:mould_lump_round}). The two-part mold for the insert also has the added benefit of enabling the manufacture of more complex inclusions (e.g. \cref{fig:mould_lump_curved}).

To manufacture the insert, we pour silicone into the two-part mold to form the top skin. In parallel, we pour silicone into the inclusion's mold. We let the silicone cure and, as with the shell, we remove the inner part of the mold but leave the cured silicone attached to the outer part, as in \cref{fig:mould_real_insert_top}. We remove the inclusion from its mold and put it into a designated hole in the base. Then, as with the shell, we pour silicone onto the base so that it is a few millimeters high, let it seep into the anchor holes, and cover it with the top part with silicone attached as in \cref{fig:insert_final_step}.

\begin{figure*}
    \centering
    \includegraphics[width=0.75\textwidth]{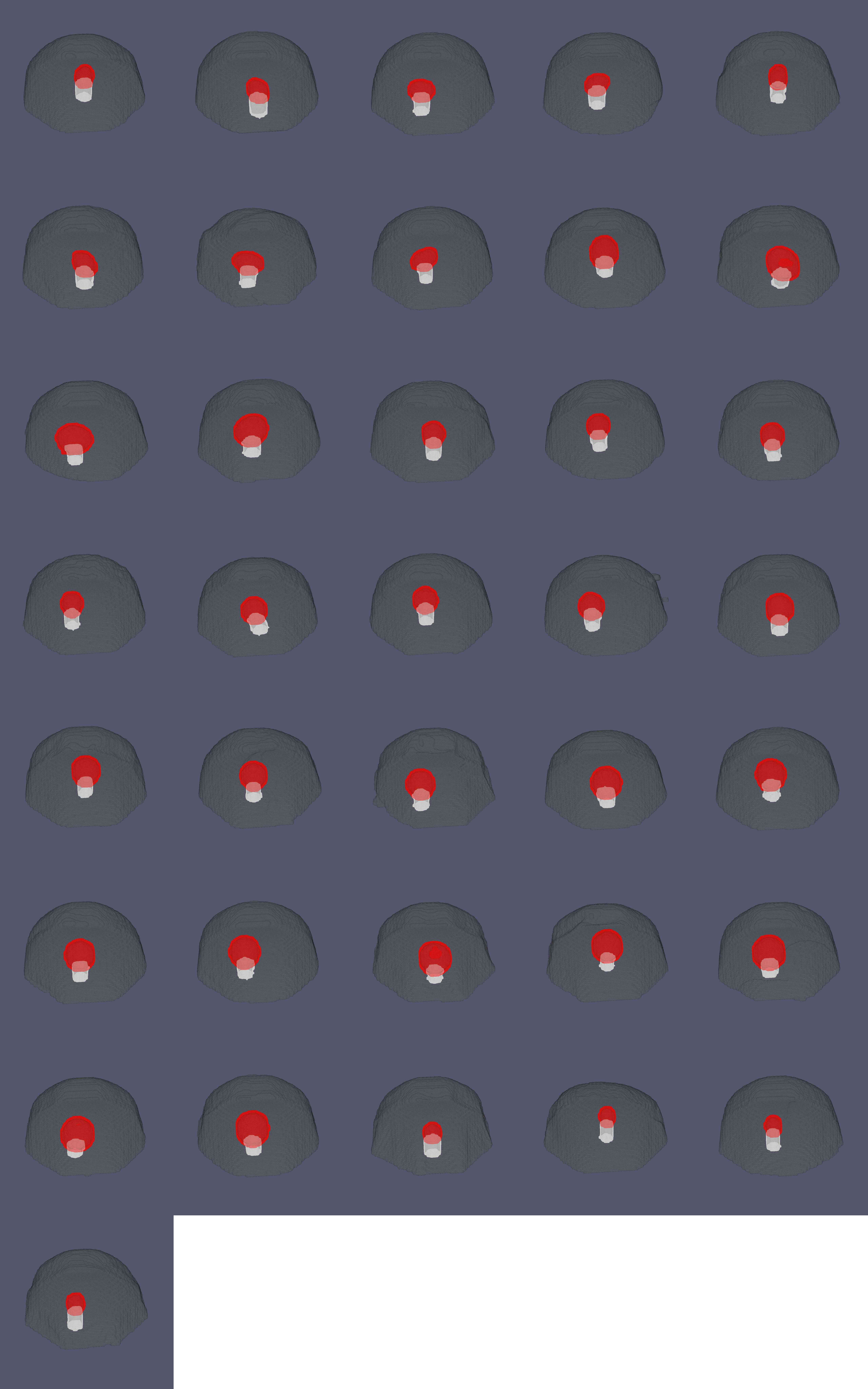}
    \caption{All our manufactured single-inclusion inserts.}
    \label{fig:all_inserts}
\end{figure*}

\begin{figure*}
    \centering
    \includegraphics[width=0.99\textwidth]{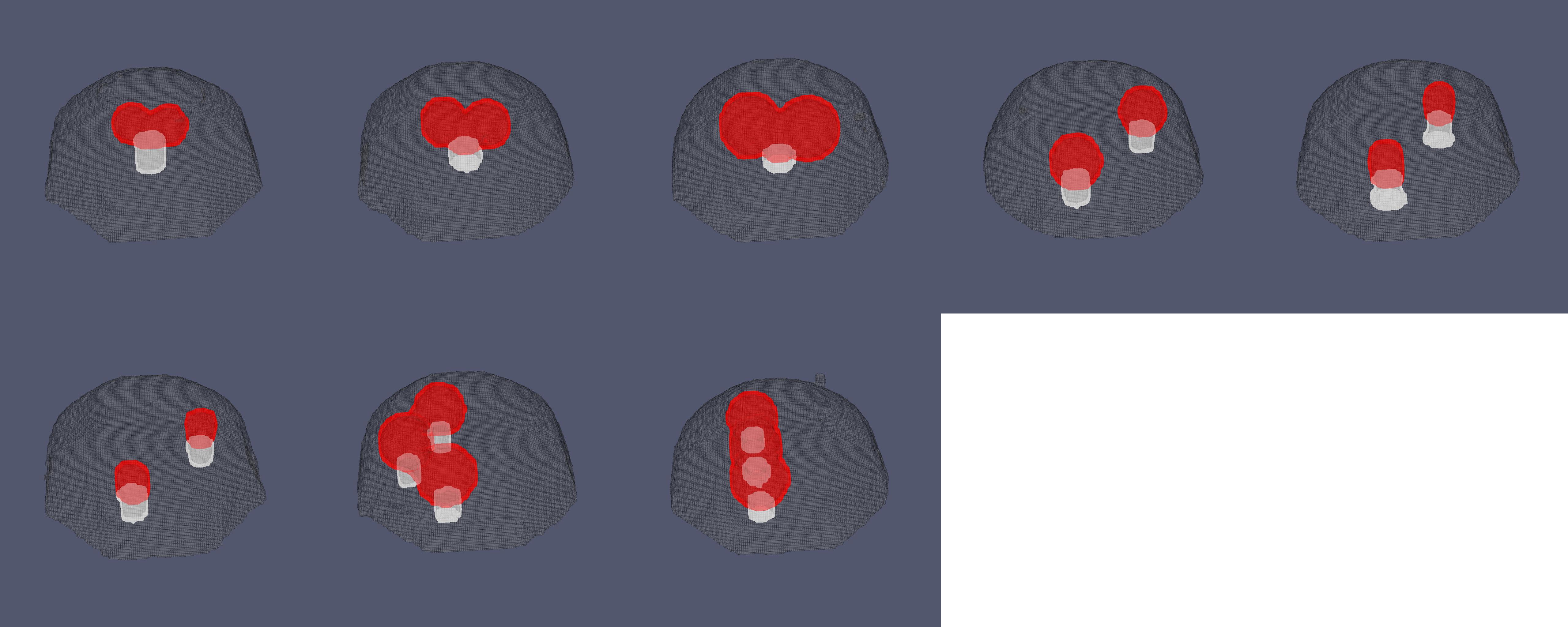}
    \caption{All our manufactured multi-inclusion inserts.}
    \label{fig:all_inserts_multi}
\end{figure*}

We have manufactured the inserts from \cite{Rimon2025ArtificialPalpation} as well as new ones (with similar inclusions), all of which are shown in \cref{fig:all_inserts}. In addition, we manufacture inserts to test the generalization of \name\, all of which are shown in \cref{fig:all_inserts_multi}. 

\begin{figure}[ht] 
    \centering
    \includegraphics[width=0.25\textwidth]{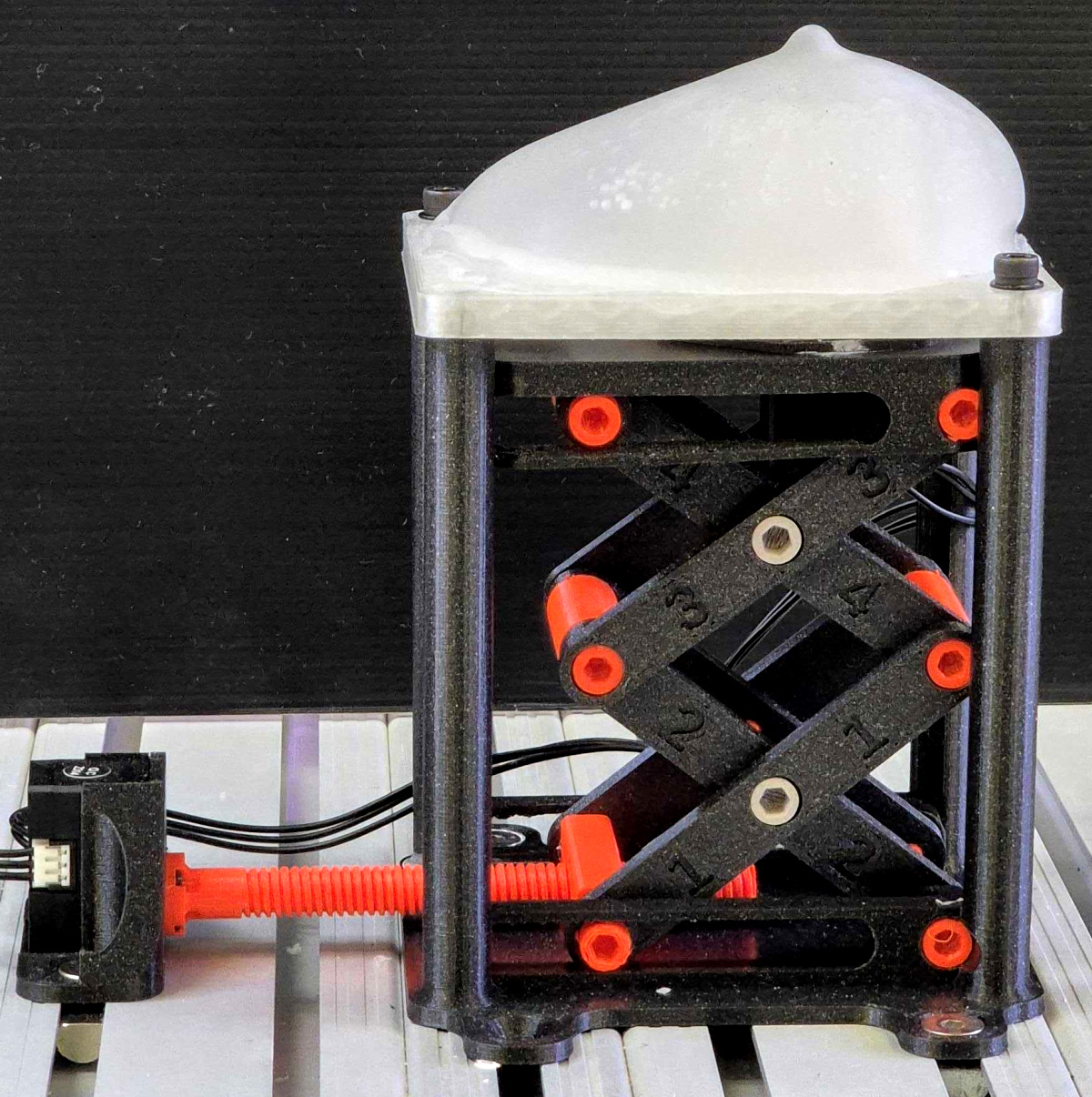}
    \caption{Scissor lift. A motor drives a screw to lift the contraption. On the top black platform, there is another motor to rotate the insert.}
    \label{fig:scissor_lift}
\end{figure}

\textbf{Lift} We tried collecting data in the same fashion as in \cite{Rimon2025ArtificialPalpation}, but found the process to be slow, since after collecting data for a single configuration we would need to remove the insert manually and put it in again. Since the base is an octagon, we can collect eight different configurations for one insert. We decided to create a mechanism to lower the insert from under the shell, rotate it, and then put it back in place (see \cref{fig:scissor_lift}).\\
The mechanism created is a motorized scissor lift based on print files for a manual lift \cite{scissor_lift}. We added a Dynamixel servo to drive the lift and another to rotate the insert. To further streamline data collection, we use 3 lifts for a single robot arm, enabling data collection without human intervention for hours on end.

\begin{figure}[ht] 
    \centering
    \includegraphics[width=0.4\textwidth]{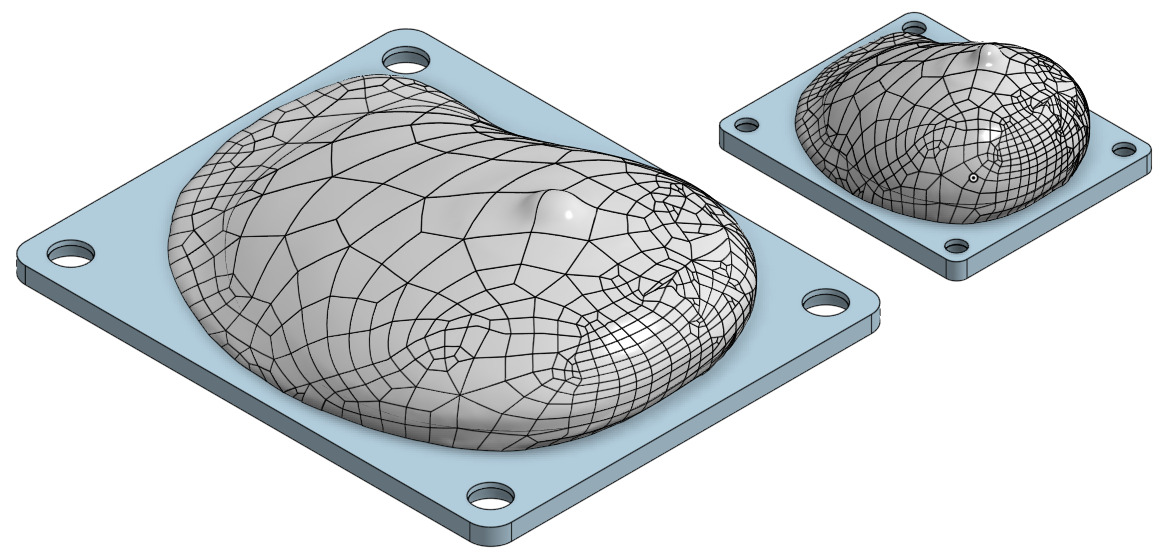}
    \caption{On the right, a regular-sized phantom. On the left, a large phantom, which is a $\times 2$ scaled version of the regular phantom in the x and y directions.}
    \label{fig:mould_big}
\end{figure}

%% file: sections/appendix/3d_gt.tex
\subsection{Ground-Truth Preparation and Alignment}
\label{sec:ground_truth_processing_and_alignment}

\begin{figure*}[ht]
    \centering
    \includegraphics[width=0.6\linewidth]{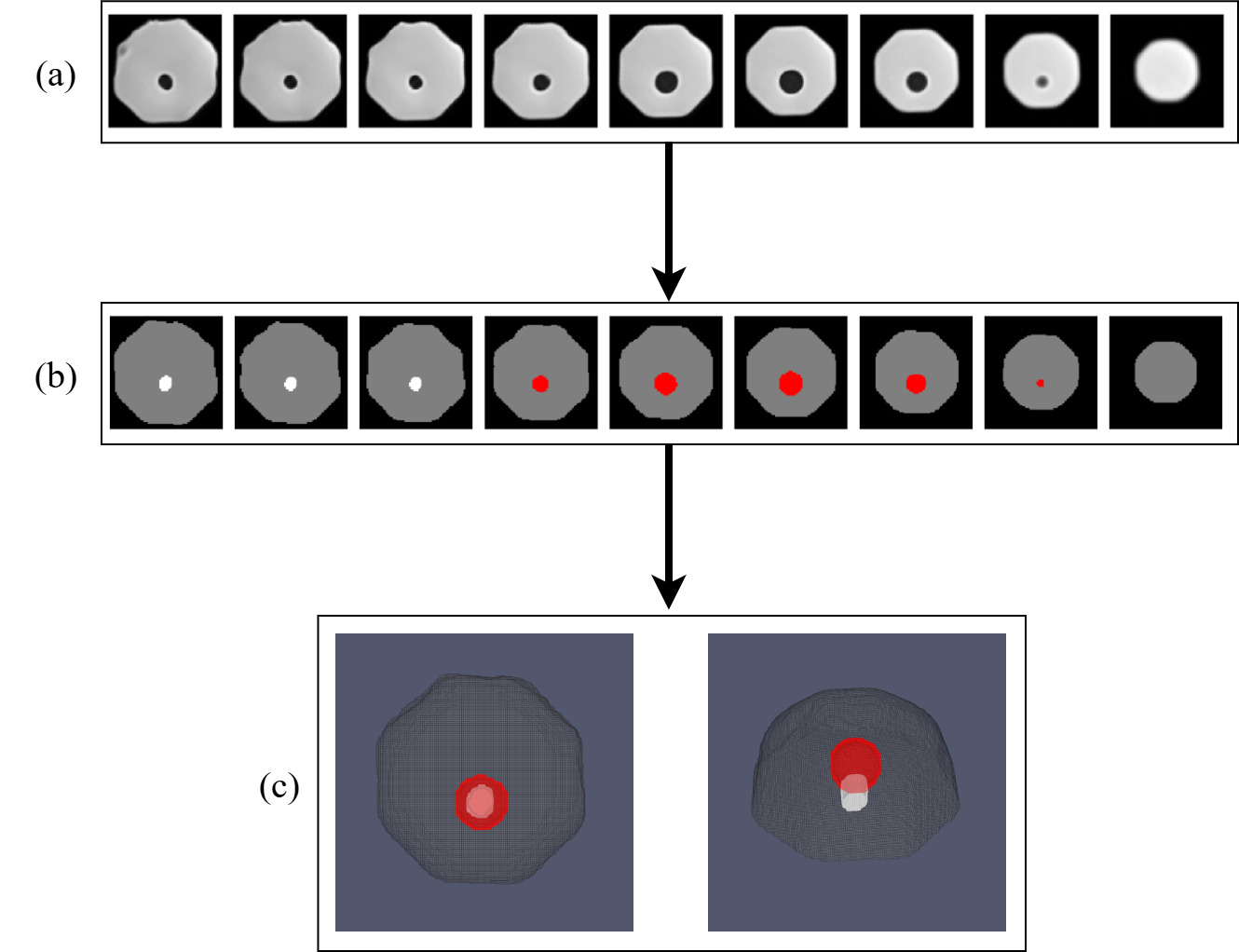}
    \caption{Overview of our pre-processing pipeline: (a) A subset of slices from the original MRI scan of a single phantom.
    (b) The corresponding processed scan with pixel-wise class annotations, where black denotes background, gray denotes the phantom insert, white denotes the pillar, and red denotes the inclusion. (c) The processed MRI scan is then sent to our custom-made visualization program, which renders the 3D scan.}
    \label{fig:GT_procedure}
\end{figure*}

As discussed in \cref{section:imaging_train_data}, each pixel in our imaging task is assigned to one of four classes: background, insert, pillar, or inclusion. To generate our ground truth, we developed an MRI processing pipeline, along with a visualization tool to interpret the ground truth and the predictions. An overview of this procedure is shown in Figure~\ref{fig:GT_procedure}.

%% file: sections/appendix/3d_metrics.tex
\subsection{Tactile Imaging Metrics} \label{sec:app_metrics}
\label{appendix:reconstruction}
\begin{figure}[ht] 
    \centering
    \includegraphics[width=0.5\textwidth]{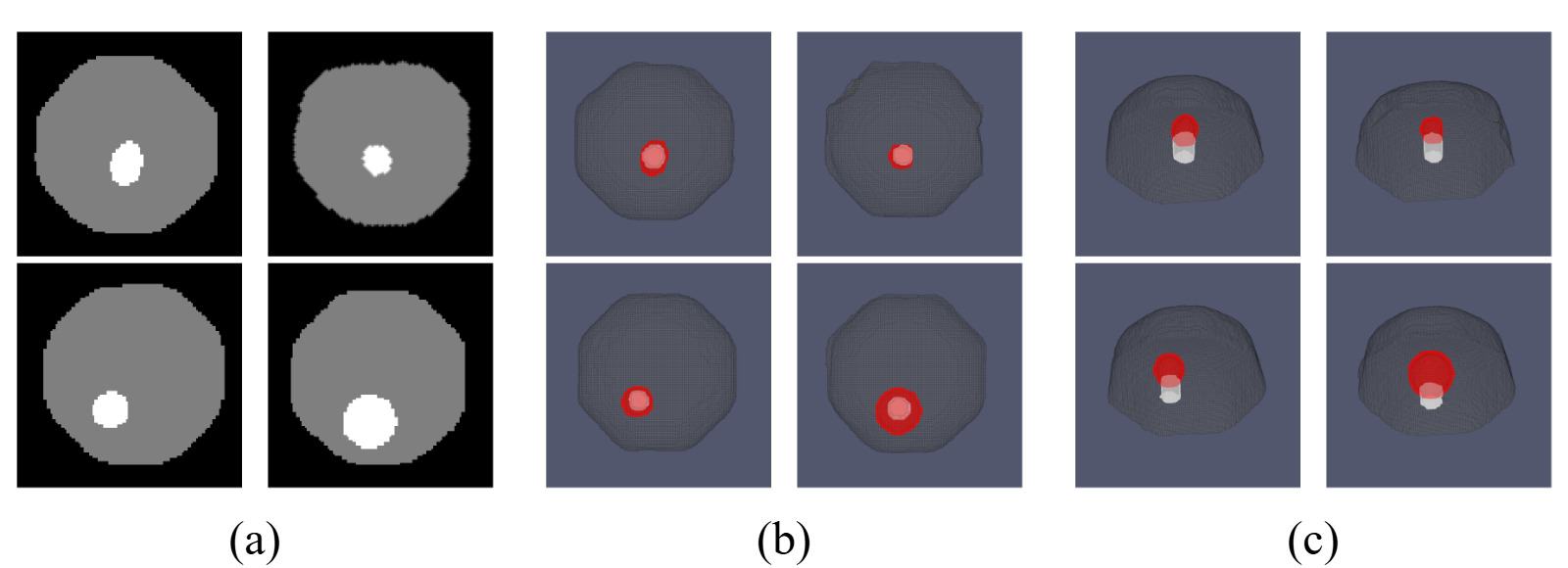}
   \caption{Comparison between 2D and 3D visualization approaches.
    (a) Representative 2D slice-based visualization as presented in \cite{Rimon2025ArtificialPalpation}.
    (b) Top-view rendering of the proposed 3D visualization.
    (c) Side-view rendering of the same 3D visualization, highlighting the vertical connectivity between the pillar and the inclusion.}
    \label{fig:gt_2d}
\end{figure}

Whereas prior work by \citet{Rimon2025ArtificialPalpation} demonstrated a proof of concept for artificial palpation using 2D tactile images, it has a key limitation: the resulting representations and tactile images are restricted to two dimensions and do not capture the 3D structure of objects, limiting information about the size, shape, and position of internal inclusions as shown in Figure~\ref{fig:gt_2d} (a). We address this limitation by generating 3D tactile images, enabling the tactile data to capture a richer and more precise representation of the objects.

In \cref{sec:exp_2d_vs_3d}, we compare predictions obtained from a full 3D reconstruction with predictions evaluated in a purely 2D setting. We report results for both 2D predictions and 3D predictions, while ensuring that evaluation is conducted under comparable conditions. In particular, metrics are computed either directly on the full 3D volumetric prediction or on a single 2D slice corresponding to the slice with the largest inclusion area. The latter follows the evaluation introduced in \citep{Rimon2025ArtificialPalpation} and enables a fair and direct comparison between 2D and 3D approaches under identical slice-level conditions.

We focus our evaluation on the inclusion class and define the following metrics, which are computed via element-wise comparison at the voxel level (for 3D volumes) and pixel level (for 2D slices):

\begin{itemize}
    \item \textbf{F1 score} - The F1 score is defined as the harmonic mean of precision and recall,
    $$\text{F1} = \frac{2 \cdot \text{precision} \cdot \text{recall}}{\text{precision} + \text{recall}} $$
    \item \textbf{Center of mass (COM) error} - The COM error is defined as the Euclidean distance between the predicted inclusion center of mass and the ground truth inclusion center of mass.
    \item \textbf{Absolute Relative Size Error (ARSE)} - Let $|p|$ denote the number of pixels classified as inclusion in the prediction and $|y|$ the number of inclusion pixels in the ground truth. The Absolute Relative Size Error is defined as
\begin{equation*}
\mathrm{ARSE} = 
\frac{\left| |p| - |y| \right|}{\frac{1}{2}\left(|p| + |y|\right)} 
\end{equation*}
    \item \textbf{Absolute Diameter Error} - We define the diameter of an inclusion as the maximum Euclidean distance, measured in millimeters, between any two pixels classified as an inclusion within the same slice. The diameter is computed for each slice independently, and the final inclusion diameter is defined as the maximum diameter over all slices. The Absolute Diameter Error is then defined as the absolute difference between the predicted diameter and the corresponding ground-truth diameter.
\end{itemize}
In Table~\ref{tab:2D_vs_3D}, we observe that predicting the full 3D structure not only improves the interpretability of the results for human observers, but also leads to consistent improvements across all 2D metrics. These findings suggest that the added geometric information captured by the 3D reconstruction provides measurable benefits beyond visualization alone.

%% file: sections/appendix/focal_loss.tex
\subsection{Imaging Class Imbalance}
\label{app:class_imb}

In \citep{Rimon2025ArtificialPalpation}, a cross-entropy loss for all the image pixels was used. This loss can be directly extended to voxel predictions. However, we found that when extending the formulation to 3D reconstruction, in particular, moving from 2D slices introduced a substantial increase in class imbalance, as the proportion of voxels belonging to the inclusion class decreased significantly relative to the background and phantom classes. This shift is quantified in Table~\ref{tab:class_imbalance}, which reports the mean and standard deviation of the number of pixels (or voxels) per class across the dataset.
\begin{table}[ht!]
  \centering
  \small
  \renewcommand{\arraystretch}{1.2}
  \begin{tabular}{|l|c|c|}
    \hline
    \textbf{Class} & \textbf{2D (\%)} & \textbf{3D (\%)} \\ \hline
    Background & $47.40 \pm 2.04$ & $50.97 \pm 2.17$ \\ \hline
    Phantom    & $49.78 \pm 2.07$ & $47.71 \pm 2.17$ \\ \hline
    Pillar     & ---                  & $0.33 \pm 0.04$ \\ \hline
    Inclusion       & $2.82 \pm 1.01$  & $0.99 \pm 0.48$ \\ \hline
  \end{tabular}
  \caption{Per-Pixel Class Distribution}
  \label{tab:class_imbalance}
\end{table}

In the presence of severe class imbalance, cross-entropy loss has been shown to struggle \cite{lin2017focal}, resulting in insufficient learning performance for minority classes. This observation motivated us to adopt a loss function that directly addresses class imbalance. In particular, we use a linear interpolation between \textbf{focal loss} and \textbf{Dice loss}, which combines the strengths of both approaches. Focal loss is designed to tackle extreme class imbalance by down-weighting easy-to-classify examples, while Dice loss optimizes the overlap between different classes, which encourages more precise boundaries.

We will now formally define our loss function. To ensure balanced gradients during optimization, we employ a dynamic scaling strategy. The total loss is defined as:
\begin{align*}
\mathcal{L} &= \lambda \cdot \mathcal{L}_{\text{Dice}} + (1 - \lambda) \cdot \eta \cdot \mathcal{L}_{\text{Focal}}, \\
\eta &= \frac{\mathcal{L}_{\text{Dice}}}{\mathcal{L}_{\text{Focal}} + \delta}, \\
\mathcal{L}_{\text{Focal}} &= - \frac{1}{HWD} \sum_{i,j,k} \sum_{c=1}^C \alpha_c y_{ijk,c} (1 - p_{ijk,c})^\gamma \log(p_{ijk,c}), \\
\mathcal{L}_{\text{Dice}} &= 1 - \frac{1}{C} \sum_{c=1}^C \frac{2 \sum_{i,j,k} p_{ijk,c} y_{ijk,c} + \epsilon}
{\sum_{i,j,k} p_{ijk,c} + \sum_{i,j,k} y_{ijk,c} + \epsilon}
\end{align*}

Here, $H$, $W$, and $D$ denote the height, width, and depth of the input volume, respectively, and $C$ represents the number of classes (4 in our case). The variable $y_{ijk,c}$ denotes the one-hot encoded ground truth (equal to 1 if the voxel at $(i,j,k)$ belongs to class $c$, and 0 otherwise), while $p_{ijk,c}$ represents the corresponding predicted softmax probability. To prevent the Dice loss from dominating the optimization, we introduce a scaling factor $\eta$, which dynamically normalizes the Focal loss magnitude to match the Dice loss at each iteration. We set $\lambda=0.5$ and the focusing parameter $\gamma = 2$. The class-balancing weights $\alpha_c$ are calculated based on the inverse frequency of each class within the batch. Finally, $\epsilon = 10^{-6}$ and $\delta = 10^{-8}$ are included for numerical stability.

%% file: sections/appendix/ssl_loss.tex
\subsection{\name\ Self-Supervised Loss}
\label{sec:app_training_details}

We adapt the force-prediction self-supervised loss from \citet{Rimon2025ArtificialPalpation} to our local prediction.

To allow for parallel computation of all particles' sequences, after filtering the sequences according to the receptive fields, we pad each local input sequence with the last measurement to a constant length $L$. In the following reconstruction procedure, we do not sample steps involving padding measurements. 

We use a mean squared error (MSE) reconstruction loss between the predicted and true forces. We uniformly subsample reconstruction steps as described in \cite{Rimon2025ArtificialPalpation} to mitigate memory complexity: 

$$
\mathcal{L}_{rec} = \frac{1}{2\nparticles\subsample \subsample'}  \sum_{n=1}^\nparticles \sum_{k=1}^\subsample 
\sum_{k'=1}^{\subsample'}
\norm{FD\left(\repr_{t_k^n}^{\lpose_{n}}, \spose_{t'^{n}_{k'}} - \lpose_{n} \right) - \force_{t'^{n}_{k'}}}^2,
$$

where FD is the Force Decoder, as described in \cref{section:architecture}, for each $n\in \left[1, \nparticles \right]$, $\left\{t^n_k\right\}_{k=1}^\subsample$ are $\subsample < L$ uniform samples without replacement from the non-padded input measurements of $\lpose_{n}$, and for each $k \in \left[1, K\right]$, $\left\{t^{\prime n}_{k'}\right\}_{k'=1}^{\subsample'}$ are also $\subsample' < L$ uniform samples. We set $\subsample= \subsample'=16$.

%% file: sections/appendix/pose_estimation_details.tex
\subsection{Pose Estimation System Technical Details}
\label{app:pose_estimation_system}

\begin{figure}
    \centering
    \includegraphics[width=1\linewidth]{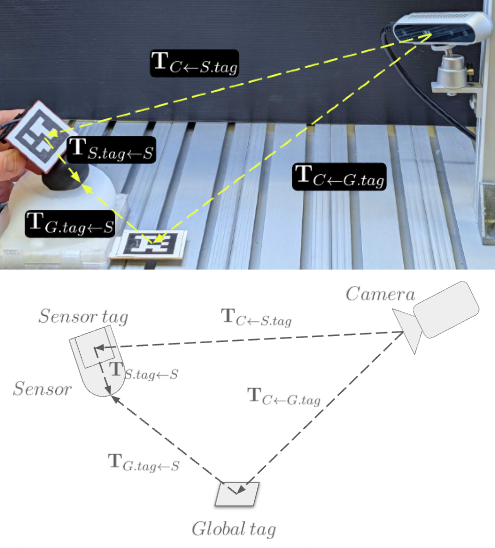}
    \caption{Pose estimation system}
    \label{fig:setup_illustration}
\end{figure}

Our setup comprises a tactile sensor (Xela uSkin) mounted on a custom 3D-printed, hand-held device. During operation, the user palpates the phantom using the sensor to acquire tactile data, while the model generates a real-time heat map localizing hard inclusions within the phantom. Accurately measuring each sample's spatial position of the sensor is required for our model.

To this end, we employ a visual tracking system using two AprilTags (each $31 \times 31\,\mathrm{mm}$): a \emph{sensor tag} mounted on the hand-held device and a \emph{global tag} placed near the phantom as a static reference. These tags are continuously monitored by a calibrated, fixed Intel RealSense 415 camera as shown in Figure~\ref{fig:setup_illustration}. Based on empirical optimization, the camera is configured to operate at 30 FPS, with a fixed exposure time of 4 ms and a sensor gain of 160 (device units), utilizing the left infrared (IR) imager (emitter disabled) for robust tag detection.

To estimate the spatial pose of the sensor relative to the phantom, we use a chain of rigid body transformations in $SE(3)$. 

Let $\mathbf{T}_{A \leftarrow B} \in SE(3)$ be a rigid transform that maps points expressed in frame $B$ into frame $A$ (i.e., $\mathbf{x}_A = \mathbf{T}_{A \leftarrow B}\,\mathbf{x}_B$ in homogeneous coordinates). We denote the transformation from the sensor tag to the sensor as $\mathbf{T}_{{S.tag} \leftarrow S} $ and obtain it via the CAD model of the device. During sampling, the camera estimates the transformations to both the sensor and global tag denoted as $\mathbf{T}_{{C} \leftarrow {S.tag}} $ and $\mathbf{T}_{{C} \leftarrow {G.tag}} $. By the identity $\mathbf{T}^{-1}_{C \leftarrow G.tag} = \mathbf{T}_{G.tag \leftarrow C} $ we calculate the sensor's pose in the global reference frame as:

$\mathbf{T}_{{G.tag} \leftarrow {S}} = \mathbf{T}_{{G.tag} \leftarrow {C} } \cdot  \mathbf{T}_{{C} \leftarrow {S.tag}} \cdot \mathbf{T}_{{S.tag} \leftarrow S}$.

\begin{figure}[th]
  \centering
  \includegraphics[width=0.9\columnwidth]{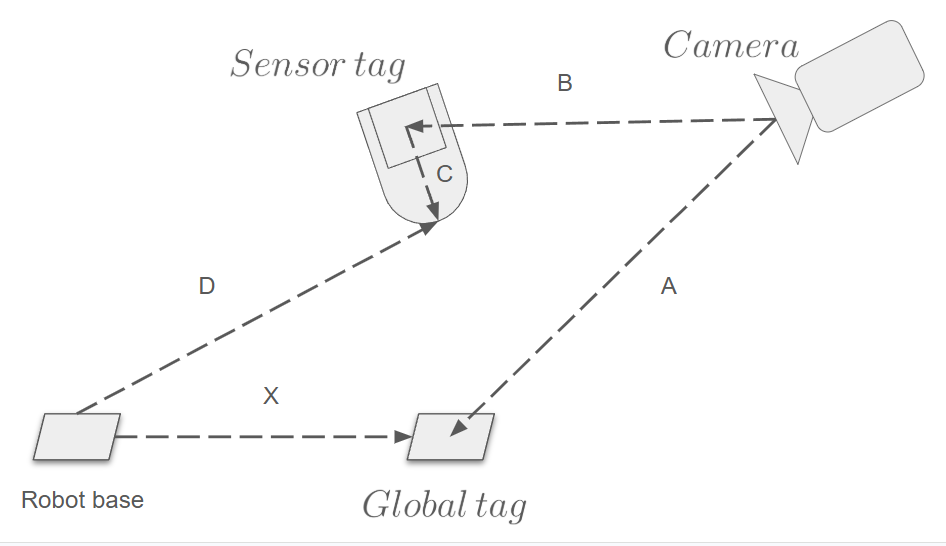}
  \caption{Calibration procedure of pose estimation to the robot's system.}
  \label{fig:calib}
\end{figure}

To calibrate the pose of the sensor from the cameras to the robot base frame, as illustrated in \cref{fig:calib}, we used a standard Kabsch algorithm to find a transformation $X$ such that $X
\cdot A^{-1} \cdot B \cdot C = D$. $A,B$ are obtained from April tags (using pupil-apriltags; camera calibration parameters obtained by realsense SDK), $C$ from the CAD of our 3D-printed rig, and $D$ from robot kinematics.

%% file: sections/appendix/pose_estimation_accuracy.tex
\subsection{Pose Estimation Accuracy}
\label{app:pose_estimation_accuracy}

To evaluate the accuracy of the system, we recorded the estimated pose during the sampling sequence and compared it against the pose from the robot's kinematics, which serves as the ground truth. This yielded two time-indexed trajectories representing the device's motion. \\ Given that the robot's reporting frequency exceeds that of our optical system, the trajectories were temporally synchronized via interpolation. Assuming the error follows a zero mean independent Gaussian distribution, we computed the empirical $\sigma$ as: $\hat{\sigma}_{positional} = \sqrt{\frac{1}{3N} \sum_{i=1}^{N} (e^2_{i,x} + e^2_{i,y} + e^2_{i,z})}$ for both positional and angular errors. The results show $\hat{\sigma}_{positional}\leq 0.45mm$ and $\hat{\sigma}_{angular}\leq 0.012\;radians$. These satisfy the accuracy requirements established in the previous section.

As the system’s error is unlikely to follow an independent Gaussian distribution in practice, we opted for an end-to-end evaluation of performance degradation. We ran the model on a sample collected using both the pose estimation and the robot’s true pose (as above) and evaluated the \name\ F1 score under each condition. The model achieved F1 scores of $67.6\pm0.0$ and $67.3\pm0.1$ when using the robot’s true pose and the estimated pose, respectively.

%% file: sections/appendix/general_data_collection.tex
\subsection{Automatic Data Collection}
\label{app:data_collection}

\begin{figure*}[ht] 
    \centering
    \includegraphics[width=0.8\textwidth]{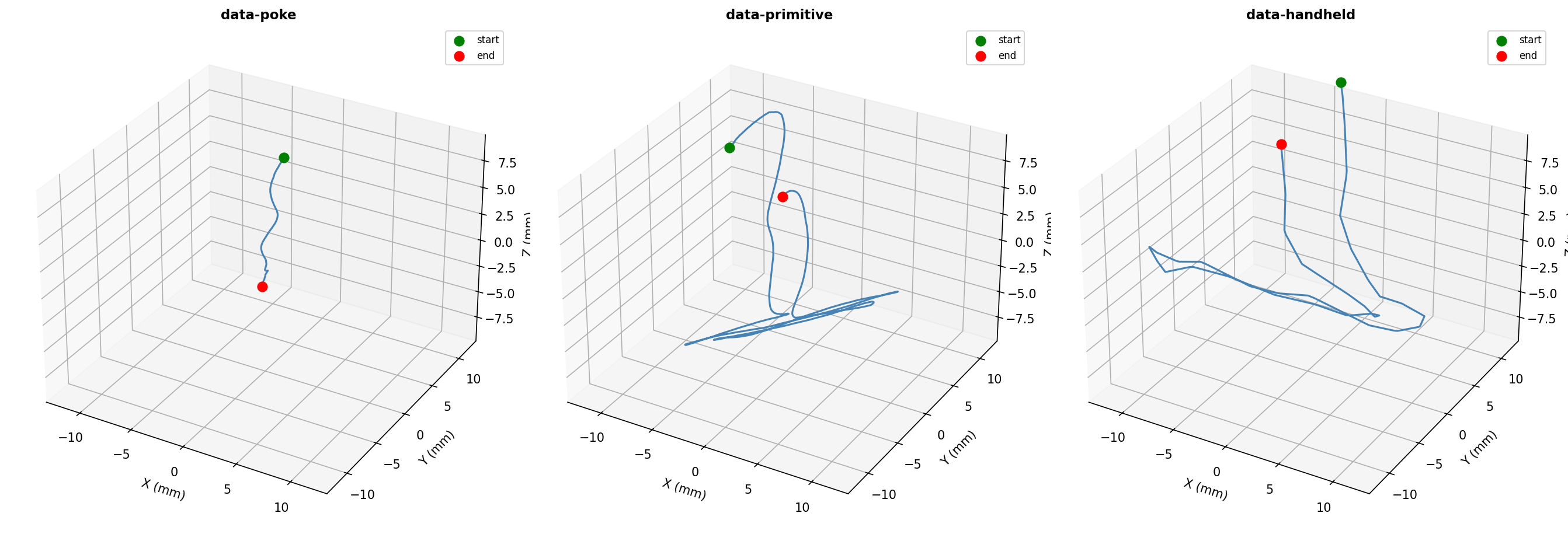}
   \caption{Comparison of a sampled trajectory from the different datasets. The primitives are more similar to the \texttt{data-handheld}, compared to \texttt{data-poke}}
    \label{fig:primitives_vis}
\end{figure*} 

We have performed our data collection using a Franka Emika Panda robot and a slightly modified panda\_py package \citep{elsner2023taming}. In order to collect a vast and diverse dataset, we implemented an automated sampling system. Insert rotation was automated using custom 3D-printed lifts equipped with rotating ramps as explained in \refapp{appendix:fabrication}. Pairing this capability with adding randomization to the sampling parameters facilitated continuous data acquisition 24/7 since we had 3 elevators, so we had sweeps of about 8 hours on 3 phantoms, and because each scan is different from the other, collecting more data on the same phantoms is still beneficial, increasing the variance of the resulting dataset. The randomness was introduced by applying spatial noise to the grid coordinates (smaller than the grid step) and stochastic adjustments to the end effector yaw angle ($0-80$ degrees). We have collected $3372$ scans with a random yaw angle and randomized positions. The scans have been performed on $8$ shells, $54$ inserts, and with $3$ Xela sensors. The data collection was done with the same controller as in \citet{Rimon2025ArtificialPalpation}.

%% file: sections/appendix/primitives.tex
\subsection{Human Primitives Based Data Collection}
\label{app:primitives}

In order to train the model to be able to work well with a hand-held sensor, we needed a way to imitate human motions. Our solution was to record several human-generated primitive motions using the robot. An operator performed several basic palpation motions on the phantom while the robot’s poses were recorded. Those primitive trajectories were then used instead of the robot pokes. While sampling each grid point, we choose a random primitive and calculate a trajectory that will originate there. Each trajectory is evaluated to ensure all points remain within the predefined boundaries of the specific phantom. If a trajectory exceeds these limits but maintains sufficient length, it is truncated; otherwise, a different motion primitive is selected. Three basic primitives were used: a poke, a back-and-forth palpation movement, and a left-and-right palpation movement. Videos of these motions are available on the website. A sampled trajectory from the primitives, compared to the \texttt{poke-data} and \texttt{poke-handheld} can be seen in \cref{fig:primitives_vis}. To run those trajectories on the robot, we used panda\_py by \citet{elsner2023taming}. We used the move\_to\_joint\_position function, which calculates detailed movements for the robot from the trajectory using the TOTG algorithm by \citet{Kunz2012TimeOptimalTG}. We have sampled a total of $1409$ full scans using the motion primitives method.

%% file: sections/appendix/handheld_technical_details.tex
\subsection{Hand-Held Technical Details}
\label{app:latency}

\textbf{System Latency:} The model-forward pass is $250ms$; other system latencies are negligible in comparison, allowing $\sim 4[Hz]$ predictions for real time interaction

\textbf{Data Collection:} \texttt{data-handheld} contains $20$ complete scans (approximately $2000$ single trajectories).